\documentclass[11pt]{article}

\usepackage[margin=1in]{geometry}
\usepackage[utf8]{inputenc}
\usepackage[T1]{fontenc}
\usepackage{lmodern}
\usepackage{amsmath,amssymb,amsthm}
\usepackage{booktabs}
\usepackage{tabularx}
\usepackage{array}
\usepackage{graphicx}
\usepackage{tikz}
\usetikzlibrary{arrows.meta,positioning,shapes.geometric,fit,calc}
\usepackage{natbib}
\usepackage{xcolor}
\usepackage{enumitem}
\usepackage{listings}
\usepackage{microtype}
\usepackage{hyperref}

\hypersetup{
    colorlinks=true,
    linkcolor=blue!55!black,
    citecolor=blue!55!black,
    urlcolor=blue!55!black,
    pdftitle={Clone-Invariant Choice Heads: Eliminating Option-Splitting Bias in Single-Pass Typed Decision Models},
    pdfauthor={Farjan Ahmmed, Izaz Ahmmed Tuhin}
}
\expandafter\def\expandafter\UrlBreaks\expandafter{\UrlBreaks\do\_}

\definecolor{calloutbg}{RGB}{248,250,252}
\definecolor{calloutborder}{RGB}{148,163,184}
\definecolor{codebg}{RGB}{249,250,251}
\definecolor{entanglefill}{RGB}{254,242,242}
\definecolor{entanglestroke}{RGB}{185,28,28}
\definecolor{cleanfill}{RGB}{239,246,255}
\definecolor{cleanstroke}{RGB}{29,78,216}
\definecolor{decouplefill}{RGB}{240,253,244}
\definecolor{decouplestroke}{RGB}{21,128,61}

\newtheorem{definition}{Definition}
\newtheorem{proposition}{Proposition}
\newtheorem{theorem}{Theorem}

\newcolumntype{L}[1]{>{\raggedright\arraybackslash}p{#1}}
\newcolumntype{Y}{>{\raggedright\arraybackslash}X}

\title{\textbf{Clone-Invariant Choice Heads: Eliminating Option-Splitting Bias in Single-Pass Typed Decision Models}}

\author{
    \textbf{Farjan Ahmmed}\\
    Department of Software Engineering, Daffodil International University\\
    Dhaka, Bangladesh\\
    \texttt{ahmmed2305341738@diu.edu.bd}
    \and
    \textbf{Izaz Ahmmed Tuhin}\\
    Lecturer, Department of Software Engineering, Daffodil International University\\
    Dhaka, Bangladesh\\
    \texttt{izaz.swe@diu.edu.bd}
}

\date{}

\begin{document}

\maketitle

\begin{center}
    \vspace{-1.2em}
    \small\textbf{Code \& Benchmarks:} \url{https://github.com/shejanahmmed/gram-choice-head}
    \vspace{0.8em}
\end{center}

\begin{abstract}
Single-pass System-1 typed decision models map context states and caller-supplied option sets to calibrated choice probabilities in one forward pass, enabling low-latency agentic routing and arbitration. Because standard Softmax readouts obey Luce's Independence of Irrelevant Alternatives (IIA), these models fail under option splitting: injecting $m=5$ paraphrased distractor clones dilutes target confidence, dropping accuracy up to 39.0\% (96.0\% to 57.0\%) and inducing up to 47.0\% Decision Flip Rate (DFR), while scalar Temperature Scaling cannot reverse IIA rank inversions. We diagnose a two-level failure hierarchy separating Level-1 Softmax partition crowding from Level-2 cross-option attention entanglement, where joint self-attention inverts 32\% of pre-readout target rankings and per-question centering collapses clone cosine similarity (0.984 to 0.193). For Level-1 crowding, we introduce the training-free \textbf{Gram-Discounted Choice Head} with \textbf{Global Corpus Mean Centering}. For Level-2 entanglement, we combine \textbf{Block-Diagonal Parallel-Mask LoRA}, a \textbf{Question-Conditioned Orthogonal Set Adapter (OSA)}, and \textbf{Disentangled Similarity-Gated Routing}. Across 1,300 queries spanning three benchmarks (\texttt{CLINC150}, \texttt{Banking77}, \texttt{OpenDecisions}) and three architectures (\texttt{Laya-421M}, \texttt{Qwen3-0.6B-RLCD}, \texttt{Jeva-4B}), our methods recover up to +24.0\% zero-shot accuracy and preserve 100\% clean baseline accuracy (88.0\%, 94.0\%), compressing cloned distractor mass from 0.4852 to 0.0457.
\end{abstract}

\section{Introduction}
\label{sec:intro}

\subsection{Typed Decision Interfaces in Single-Pass AI Systems}
\label{sec:intro_typed}

Production AI agents, tool routers, and safety guardrails routinely execute discrete choices under tight millisecond latency budgets. Rather than invoking slow autoregressive decoding to generate free-form JSON text and parsing it post hoc, recent systems deploy single-pass \textbf{System-1 typed decision models}---including bidirectional encoders (\texttt{Laya-421M} \citep{laya2026}), prefix-cached causal decoders (\texttt{Qwen3-0.6B-RLCD} \citep{jeff2026}), and cross-candidate matrix stacks (\texttt{Jeva-4B} \citep{jeva2026}). Given a context state $s \in \mathcal{S}$ and a caller-supplied option set $\mathcal{O} = \{o_1, \dots, o_K\}$, a typed decision model evaluates all $K$ candidates in one forward pass and emits a categorical distribution $P(o_i \mid s, \mathcal{O})$. Downstream control loops rely on a foundational operational contract: $P(o_i \mid s, \mathcal{O})$ must represent a calibrated choice probability over semantic intents that remains invariant when tool registries, retrieval pipelines, or planners introduce redundant paraphrases or fine-grained sub-splits into $\mathcal{O}$.

\subsection{The Option-Splitting Vulnerability}
\label{sec:intro_vuln}

Despite differences in sequence encoding, existing typed decision models terminate in an unweighted categorical Softmax readout over scalar logits $z_i \in \mathbb{R}$: $P(o_i \mid s, \mathcal{O}) = \exp(z_i) / \sum_{j=1}^K \exp(z_j)$. By construction, Softmax enforces Luce's Choice Axiom \citep{luce1959}, or the \emph{Independence of Irrelevant Alternatives} (IIA \citep{mcfadden1974}), where relative odds $P(o_i)/P(o_j) = \exp(z_i - z_j)$ ignore candidate redundancy. In real routing pipelines, tool schemas contain overlapping API wrappers, retrieval routers surface paraphrased policy rules, and taxonomies split a single wrong intent into multiple sub-categories---the neural analog of Debreu's \emph{Red Bus / Blue Bus} paradox \citep{debreu1960,tversky1972}. Splitting a distractor steals probability mass from the true target $o^*$ and flips top-1 decisions.

\vspace{0.6em}
\noindent\fcolorbox{calloutborder}{calloutbg}{%
\parbox{\dimexpr\linewidth-2\fboxsep-2\fboxrule\relax}{%
\small
\textbf{Box 1: Empirical Option-Splitting Inversion on \texttt{Laya-421M} (\texttt{triage\_003\_exp01})}
\begin{itemize}[leftmargin=1.2em,itemsep=2pt,topsep=4pt,parsep=0pt]
    \item \textbf{Context State ($s$):} \emph{``Urgent notice: User message: `Can you tell me what the annual percentage rate is on the platinum rewards credit card and whether there is an annual fee?' Please expedite resolution immediately.''}
    \item \textbf{Instruction:} \emph{``What is the primary customer intent?''}
    \item \textbf{Clean Options ($\mathcal{O}_{m=0}, K=4$):} \texttt{card\_terms\_inquiry} ($o^*$: \emph{``Inquiries regarding credit card APR, interest rates, rewards, and annual fees''}), \texttt{wire\_transfer\_request}, \texttt{mortgage\_application}, and \texttt{atm\_locator} ($o_{\text{dist}}$: \emph{``Finding nearby physical branch ATMs and cash deposit locations''}).
    \item \textbf{Clean Prediction ($m=0$):} \textbf{$P(o^* = \texttt{card\_terms\_inquiry}) = 0.9111$} vs.\ \textbf{$P(\texttt{atm\_locator}) = 0.0295$}.
    \item \textbf{Cloned Options ($\mathcal{O}_{m=5}, K=9$):} Injecting $m=5$ paraphrased clones of \texttt{atm\_locator} (\texttt{atm\_locator\_clone1..5}, e.g., \emph{``Locating the nearest branch ATM and branch office address''}, \emph{``Finding cash deposit machines and physical bank ATM kiosks''}).
    \item \textbf{Corrupted Prediction ($m=5$):} Target probability collapses from \textbf{$0.9111 \to 0.1688$}, while the \texttt{atm\_locator} cluster absorbs \textbf{$\sum_{c \in \mathcal{C}_{\text{dist}}} P(c) = 0.8172$} ($P(\texttt{atm\_locator}) = 0.5800$), flipping the top-1 prediction to an unrelated ATM search intent.
\end{itemize}
}}
\vspace{0.6em}

\begin{figure}[t]
    \centering
    \includegraphics[width=\textwidth]{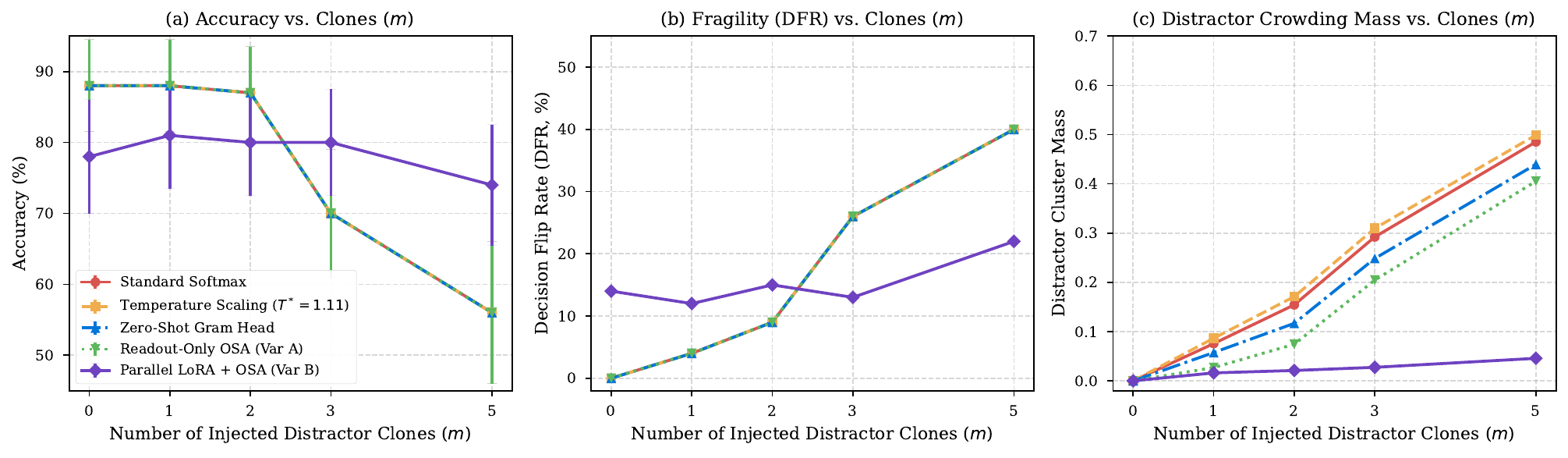}
    \caption{Degradation under distractor paraphrase cloning ($m \in \{0,1,2,3,5\}$) on the \texttt{Laya-421M} Controlled Stress Test ($N=100$ paired queries per slice). Panel (a): Top-1 accuracy; Panel (b): Decision Flip Rate (DFR); Panel (c): Distractor cluster mass.}
    \label{fig:degradation}
\end{figure}

As shown in \textbf{Figure~\ref{fig:degradation}}, this collapse is systematic across our 900-query Controlled Stress Test. In Panel (a), as distractor clone count grows from $m=0$ to $m=5$, \texttt{Laya-421M} accuracy falls from 88.0\% \texttt{[81.0, 94.0]} to 56.0\% \texttt{[46.0, 66.0]}. In Panel (b), the Decision Flip Rate (DFR) relative to clean $m=0$ predictions surges from 0.0\% to 40.0\% \texttt{[30.0, 50.0]}. In Panel (c), distractor cluster mass rises to 0.4852 (a $9.1\times$ increase over the 0.0534 uncloned single-distractor baseline, where Figure~\ref{fig:degradation}c displays 0.0000 at $m=0$ because cluster indices are empty when no clones are present), whereas our decoupled architecture (\texttt{Parallel LoRA + OSA}, Variant B) holds accuracy at 74.0\%, restricts DFR to 22.0\%, and compresses cluster mass by $10.6\times$ to 0.0457.

\subsection{Core Structural Insight: Two-Level Failure Hierarchy}
\label{sec:intro_hierarchy}

Standard post-hoc calibration cannot fix option-splitting bias: scalar Temperature Scaling ($z_i / T$, $T > 0$ \citep{guo2017calibration}) is strictly monotonic, preserving logit ordering ($\operatorname{argmax}_i z_i / T = \operatorname{argmax}_i z_i$) and leaving $m=5$ accuracy (56.0\%) and DFR (40.0\%) unchanged. We show that option splitting breaks typed decision models at \textbf{two distinct levels}:

\begin{enumerate}[leftmargin=1.5em]
    \item \textbf{Level-1 Readout Crowding (Universal):} Even when candidate representations $\mathbf{e}_i \in \mathbb{R}^d$ and logits $z_i$ are scored independently, duplicating a distractor $m$ times adds $m$ exponential terms to the Softmax partition sum, diluting target confidence at rate $\mathcal{O}(1/m)$. Moreover, unweighted per-question centering ($\mathbf{e}_i - \frac{1}{K}\sum_j \mathbf{e}_j$) suffers from a \emph{self-subtraction pathology}: when $m=5$ clones form $6/9$ of $\mathcal{O}$, the sample mean aligns with the clone cluster and collapses intra-clone cosine similarity from 0.984 to 0.193.
    \item \textbf{Level-2 Internal Cross-Option Entanglement (Architecture-Specific):} In bidirectional encoders (\texttt{Laya-421M}) and cross-option matrix stacks (\texttt{Jeva-4B}), candidate tokens attend to one another across transformer layers. Self-attention across $m+1$ near-duplicate distractors both amplifies distractor logits---inverting 32/100 clean-correct target rankings at $m=5$ (mean deficit $\Delta z = 1.19\text{ nats}$, up to $2.91\text{ nats}$)---and entangles the target representation itself (mean effective crowd size $c^* = 6.60 > c_{\text{clone}} = 5.21$), neutralizing post-readout discounting.
\end{enumerate}

\subsection{Summary of Contributions}
\label{sec:intro_contrib}

\begin{itemize}[leftmargin=1.5em]
    \item \textbf{Clone-Invariance Axioms \& Training-Free Gram-Discounted Readout (Section~\ref{sec:prob}, Theorem~\ref{thm:clone_inv}):} Formalize exact clone invariance and equivalence-class mass conservation; prove Softmax's $\mathcal{O}(1/m)$ dilution and Temperature Scaling's rank invariance (Propositions~\ref{prop:temp}--\ref{prop:dilution}); introduce \texttt{GramDiscountedChoiceHead} ($\tilde{z}_i = z_i / T - \log c_i$, $c_i = \sum_j S_{ij}^\gamma$) paired with \textbf{Global Corpus Mean Centering} (0.931 clone vs.\ 0.463 distinct similarity), recovering up to +24.0\% zero-shot accuracy (63.0\% $\to$ 87.0\% at $m=3$; 57.0\% $\to$ 74.0\% at $m=5$) on \texttt{Qwen3-0.6B-RLCD}.
    \item \textbf{Mechanistic Diagnosis \& Level-2 Structural Decoupling (Section~\ref{sec:arch}, Eqs.~\eqref{eq:margin_bound}--\eqref{eq:gate_sim}):} Derive the exact margin inversion bound ($\Delta z < \log(m+1)$) explaining why readout-only heads fail on joint encoders, and introduce \textbf{Block-Diagonal Parallel-Mask LoRA + Question-Conditioned Orthogonal Set Adapter (OSA)} (330,881 trainable parameters) with clean-teacher KL anchoring and \textbf{Disentangled Similarity-Gated Routing}.
    \item \textbf{Controlled \& Cross-Architecture Benchmark Suite (Section~\ref{sec:eval}):} Construct and release 1,300 paired evaluation queries across three benchmarks (\texttt{CLINC150} Stress Test, \texttt{Banking77}, \texttt{OpenDecisions}), three perturbation regimes (distractor paraphrasing, taxonomic sub-splitting, positive-target symmetry), and three model families (\texttt{Laya-421M}, \texttt{Qwen3-0.6B-RLCD}, \texttt{Jeva-4B}).
    \item \textbf{Empirical Verification \& Practitioner Deployment Rules (Sections~\ref{sec:results}--\ref{sec:guidance}, Tables~\ref{tab:benchmarks}--\ref{tab:deployment}, Figures~\ref{fig:degradation}--\ref{fig:reliability_ablation}):} Demonstrate zero clean-accuracy regression (88.0\% on Stress Test; 94.0\% on \texttt{Banking77}), +18.0\% accuracy recovery at $m=5$ (56.0\% $\to$ 74.0\%), $10.6\times$ distractor mass suppression (0.4852 $\to$ 0.0457), tight 3-seed stability ($72.00 \pm 2.65\%$), and boundary diagnostics across low-competence and matrix-stack regimes.
\end{itemize}

\section{Problem Formulation \& Level-1 Gram-Discounted Readout}
\label{sec:prob}

\subsection{Typed Decision Interface \& Calibration Invariance}
\label{sec:prob_interface}

Consider a single-pass typed decision model $f_\phi$ parameterized by weights $\phi$ \citep{ndmblock2026,laya2026,jeff2026}. Each query specifies a context state $s \in \mathcal{S}$ and a caller-supplied candidate option set $\mathcal{O} = \{o_1, \dots, o_K\}$ ($K \ge 2$) containing a unique ground-truth target option $o^* \in \mathcal{O}$. In one forward pass, the model maps $(s, \mathcal{O})$ to option embeddings $\mathbf{e}_i \in \mathbb{R}^d$ and pre-activation scalar logits $z_i = g_\phi(s, o_i, \mathcal{O}) \in \mathbb{R}$ for $i \in \{1, \dots, K\}$, inducing the categorical Softmax distribution:
\begin{equation}
    P_{\text{soft}}(o_i \mid s, \mathcal{O}) = \frac{\exp(z_i)}{\sum_{j=1}^K \exp(z_j)}.
    \label{eq:softmax}
\end{equation}

Equation~\eqref{eq:softmax} satisfies Luce's Choice Axiom \citep{luce1959}, or the \emph{Independence of Irrelevant Alternatives} (IIA \citep{mcfadden1974}): for any candidates $o_i, o_j \in \mathcal{O}$, the relative odds $P_{\text{soft}}(o_i \mid s, \mathcal{O}) / P_{\text{soft}}(o_j \mid s, \mathcal{O}) = \exp(z_i - z_j)$ depend strictly on $z_i - z_j$, remaining invariant to candidate multiplicity in $\mathcal{O}$. Consequently, post-hoc scalar temperature calibration cannot prevent option-splitting inversions.

\begin{proposition}[Temperature Scaling Rank Invariance]
    \label{prop:temp}
    For any logit vector $z \in \mathbb{R}^K$ and positive scalar temperature $T > 0$ \citep{guo2017calibration}, rescaling $\tilde{z}_i = z_i / T$ is strictly order-preserving ($\tilde{z}_i > \tilde{z}_j \iff z_i > z_j$), leaving $\operatorname{argmax}_{i} P(o_i \mid s, \mathcal{O})$, top-1 accuracy, and Decision Flip Rate (DFR) invariant under option splitting.
\end{proposition}

\subsection{Clone-Invariance Axioms \& Softmax Impossibility}
\label{sec:prob_axioms}

Consider a clean candidate set $\mathcal{O}^{(0)} = \{o_1, \dots, o_K\}$ of mutually distinct options. Injecting $m \ge 1$ duplicates of distractor $o_K \ne o^*$ yields an expanded option set $\mathcal{O}^{(m)} = \mathcal{O}^{(0)} \cup \{o_{K+1}, \dots, o_{K+m}\}$ of cardinality $K + m$, partitioning the duplicates into an equivalence class $\mathcal{C} = \{o_K, o_{K+1}, \dots, o_{K+m}\}$ ($|\mathcal{C}| = m + 1$).

\begin{definition}[Exact Clone Invariance \& Equivalence-Class Mass Conservation]
    \label{def:clone_inv}
    A choice readout $P$ satisfies exact clone invariance if, whenever option $o_K \in \mathcal{O}^{(0)}$ is duplicated into an equivalence class $\mathcal{C} \subseteq \mathcal{O}^{(m)}$ of $m + 1$ identical copies ($z_c = z_K, \mathbf{e}_c = \mathbf{e}_K$ for all $c \in \mathcal{C}$), the distribution $P^{(m)}(\cdot \mid s, \mathcal{O}^{(m)})$ satisfies for every $m \ge 1$:
    \begin{enumerate}[leftmargin=1.5em,itemsep=1pt,topsep=2pt]
        \item Non-Cloned Probability Invariance: $P^{(m)}(o_i \mid s, \mathcal{O}^{(m)}) = P^{(0)}(o_i \mid s, \mathcal{O}^{(0)})$ for all $o_i \notin \mathcal{C}$ (including $o^*$).
        \item Equivalence-Class Mass Conservation: $\sum_{c \in \mathcal{C}} P^{(m)}(c \mid s, \mathcal{O}^{(m)}) = P^{(0)}(o_K \mid s, \mathcal{O}^{(0)})$.
        \item Within-Class Symmetry: $P^{(m)}(c \mid s, \mathcal{O}^{(m)}) = \frac{1}{m+1} P^{(0)}(o_K \mid s, \mathcal{O}^{(0)})$ for all $c \in \mathcal{C}$.
    \end{enumerate}
\end{definition}

Permutation-equivariant set models such as Deep Sets \citep{zaheer2017deepsets} and Set Transformers \citep{lee2019settransformer} ensure $P(\pi(\mathcal{O})) = \pi(P(\mathcal{O}))$ for any permutation $\pi$, but do not satisfy Definition~\ref{def:clone_inv} when candidate cardinality grows from $K$ to $K+m$. Even when pre-activation logits remain uncorrupted ($z_c = z_K$), standard Softmax suffers unbounded target dilution.

\begin{proposition}[Softmax Dilution Limit]
    \label{prop:dilution}
    Suppose option logits $z_1, \dots, z_K$ remain invariant under duplication ($z_c = z_K$ for all $c \in \mathcal{C}$). Under standard Softmax \eqref{eq:softmax},
    \begin{equation}
        P_{\text{soft}}^{(m)}(o^* \mid s, \mathcal{O}^{(m)}) = \frac{\exp(z^*)}{\sum_{j=1}^{K-1} \exp(z_j) + (m+1)\exp(z_K)} = \mathcal{O}(1/m) \xrightarrow{m \to \infty} 0,
        \label{eq:dilution}
    \end{equation}
    while total distractor cluster mass $\sum_{c \in \mathcal{C}} P_{\text{soft}}^{(m)}(c \mid s, \mathcal{O}^{(m)}) \to 1$, causing a cluster mass inversion ($\sum_{c \in \mathcal{C}} P_{\text{soft}}^{(m)}(c) > P_{\text{soft}}^{(m)}(o^*)$) whenever $m > \exp(z^* - z_K) - 1$ (derivation in Appendix~\ref{app:proofs}).
\end{proposition}

\subsection{The Gram-Discounted Choice Head (Level-1 Intervention)}
\label{sec:prob_gram}

To enforce Definition~\ref{def:clone_inv} without modifying backbone weights or applying non-differentiable hard pruning, we estimate each option's effective crowd multiplicity from candidate representation geometry at readout.

\begin{definition}[Thresholded Similarity Gram Matrix \& Effective Crowd Size]
    \label{def:gram}
    Given option embeddings $\mathbf{e}_1, \dots, \mathbf{e}_K \in \mathbb{R}^d$ and centering reference $\boldsymbol{\mu} \in \mathbb{R}^d$, let $\tilde{\mathbf{e}}_i = \mathbf{e}_i - \boldsymbol{\mu}$ and $\hat{\mathbf{e}}_i = \tilde{\mathbf{e}}_i / \max(\|\tilde{\mathbf{e}}_i\|_2, \epsilon)$ ($\epsilon = 10^{-8}$). For similarity threshold $\tau_{\text{sim}} \in (0, 1)$, define the thresholded Gram matrix $S \in [0, 1]^{K \times K}$ with diagonal $S_{ii} = 1$ and off-diagonal entries:
    \begin{equation}
        S_{ij} = \mathbb{I}\!\left[\hat{\mathbf{e}}_i^\top \hat{\mathbf{e}}_j \ge \tau_{\text{sim}}\right] \cdot \max\!\left(0, \hat{\mathbf{e}}_i^\top \hat{\mathbf{e}}_j\right), \quad i \ne j.
        \label{eq:gram}
    \end{equation}
    For sharpness exponent $\gamma \ge 1$, define each option's effective crowd size $c_i = \sum_{j=1}^K S_{ij}^\gamma \ge 1$. At temperature $T > 0$, the Gram-Discounted Choice Head computes discounted logits $\tilde{z}_i$ and choice probabilities $P_{\text{Gram}}$ via:
    \begin{equation}
        \tilde{z}_i = \frac{z_i}{T} - \log c_i, \qquad P_{\text{Gram}}(o_i \mid s, \mathcal{O}) = \frac{\exp(\tilde{z}_i)}{\sum_{j=1}^K \exp(\tilde{z}_j)} = \frac{c_i^{-1}\exp(z_i / T)}{\sum_{j=1}^K c_j^{-1}\exp(z_j / T)}.
        \label{eq:gram_head}
    \end{equation}
\end{definition}

Temperature scaling $z_i / T$ must precede crowd subtraction $-\log c_i$: applying temperature after discounting would yield $\exp((z_i - \log c_i)/T) = c_i^{-1/T}\exp(z_i/T)$, violating clone invariance whenever $T \ne 1$. In batched inputs with padding mask $v \in \{0, 1\}^K$, padded slots are isolated via $S_{ij} \leftarrow v_i v_j S_{ij}$ and $\tilde{z}_i = -10^4$, ensuring exact padding invariance.

\begin{theorem}[Exact Clone Invariance Guarantee]
    \label{thm:clone_inv}
    Assume $o_K$ is duplicated $m \ge 1$ times with exact representation equality ($\hat{\mathbf{e}}_c = \hat{\mathbf{e}}_K$ and $z_c = z_K$ for all $c \in \mathcal{C}$) while distinct options satisfy $\hat{\mathbf{e}}_i^\top \hat{\mathbf{e}}_j < \tau_{\text{sim}}$ for all $i < K$ and $j \ne i$. Then $S_{cc'} = 1$ for all $c, c' \in \mathcal{C}$ and $S_{ic} = 0$ for all $i < K$, yielding $c_c = m + 1$ for every clone and $c_i = 1$ for every distinct option. Substituting $c_c^{-1} = (m+1)^{-1}$ into Eq.~\eqref{eq:gram_head} yields $\sum_{c \in \mathcal{C}} c_c^{-1}\exp(z_c / T) = \exp(z_K / T)$, canceling the $(m+1)$ multiplicity factor in the partition sum identically for any $T > 0$ and $\gamma \ge 1$, and satisfying all three axioms of Definition~\ref{def:clone_inv} (complete proof in Appendix~\ref{app:proofs}; verified numerically to $<10^{-6}$ tolerance in Appendix~\ref{app:tests}).
\end{theorem}

While Theorem~\ref{thm:clone_inv} analyzes exact vector equality ($\cos = 1.0$), real semantic paraphrases under Global Corpus Mean Centering exhibit mean centered cosine similarity 0.931 compared to 0.463 for distinct options (Table~\ref{tab:centering}). Setting $\tau_{\text{sim}} = 0.50$ truncates background cross-talk among distinct options ($0.463 < 0.50$), while $\gamma > 1$ (e.g., $\gamma = 2.0$ or $4.0$) attenuates near-threshold sub-category overlap and applies smooth logarithmic discounting to high-similarity paraphrase clusters ($0.931^2 = 0.867$) without non-differentiable hard pruning. The readout module defaults are $\tau_{\text{sim}} = 0.50$ and $\gamma = 2.0$. The milestone-2 validation grid on \texttt{Laya-421M} ($\gamma \in \{1, 2, 4\}$, $\tau_{\text{sim}} \in \{0, 0.3, 0.5, 0.7\}$) selects $\gamma = 1.0$, $\tau_{\text{sim}} = 0.30$ (57.0\% accuracy at $m=5$), but the grid spans only 55.0--57.0\%. The evaluation runs reported in Section~\ref{sec:results} fix their own settings, listed in Section~\ref{sec:eval_protocol}.

\subsection{Representation Geometry \& Global Corpus Mean Centering}
\label{sec:prob_centering}

Contextual transformer embeddings cluster inside a narrow anisotropic cone in $\mathbb{R}^d$ \citep{ethayarajh2019anisotropy}. On 100 $m=5$ paraphrase-cloned queries in \texttt{Laya-421M} (Table~\ref{tab:centering}), raw uncentered option vectors ($\boldsymbol{\mu} = \mathbf{0}$) yield an intra-clone cosine similarity of 0.984, but genuinely distinct options exhibit an inflated mean cosine similarity of 0.796---exceeding $\tau_{\text{sim}} = 0.50$ and triggering false crowd penalties ($c_i > 1$) even when $m = 0$.

Subtracting the per-question unweighted sample mean $\bar{\mathbf{e}}_{\mathcal{O}} = \frac{1}{K+m}\sum_{j=1}^{K+m} \mathbf{e}_j$ fails under option splitting due to a \textbf{self-subtraction pathology}. When $m=5$ clones are injected into a 4-option query ($K+m = 9$), the duplicate cluster $\mathcal{C}$ accounts for $\frac{m+1}{K+m} = \frac{6}{9} = \frac{2}{3}$ of the sample mean:
\begin{equation}
    \bar{\mathbf{e}}_{\mathcal{O}} = \frac{1}{K+m}\sum_{o_j \notin \mathcal{C}} \mathbf{e}_j + \frac{m+1}{K+m}\bar{\mathbf{e}}_{\mathcal{C}}, \qquad \bar{\mathbf{e}}_{\mathcal{C}} = \frac{1}{m+1}\sum_{c \in \mathcal{C}} \mathbf{e}_c.
    \label{eq:sample_mean}
\end{equation}
As $m$ grows, $\bar{\mathbf{e}}_{\mathcal{O}}$ is pulled toward the duplicate cluster centroid $\bar{\mathbf{e}}_{\mathcal{C}}$, shifting the coordinate origin inside the clone cluster so small lexical variations among clones point in opposing directions---collapsing mean intra-clone cosine similarity from 0.984 to 0.193 (and distinct similarity to $-0.287$, Table~\ref{tab:centering}; derived in Appendix~\ref{app:proofs}).

To remove anisotropic cone bias without shifting the origin under runtime duplication, we introduce \textbf{Global Corpus Mean Centering}:
\begin{equation}
    \tilde{\mathbf{e}}_i = \mathbf{e}_i - \boldsymbol{\mu}_{\text{global}}, \qquad \boldsymbol{\mu}_{\text{global}} = \frac{1}{|\mathcal{D}_{\text{clean}}|}\sum_{o \in \mathcal{D}_{\text{clean}}} \mathbf{e}(o),
    \label{eq:global_mean}
\end{equation}
where $\boldsymbol{\mu}_{\text{global}} \in \mathbb{R}^d$ is estimated strictly once on clean ($m=0$) training options. Because $\boldsymbol{\mu}_{\text{global}}$ is independent of the runtime option set $\mathcal{O}^{(m)}$, injecting $m$ clones never moves the coordinate origin, restoring clean geometric separation (0.931 intra-clone vs.\ 0.463 distinct similarity at $\tau_{\text{sim}} = 0.50$).

\section{Architectural Failure Taxonomy \& Level-2 Structural Decoupling}
\label{sec:arch}

\subsection{Three Architectural Families of System-1 Decision Models}
\label{sec:arch_families}

Whether the training-free \texttt{GramDiscountedChoiceHead} (Section~\ref{sec:prob_gram}) suffices alone depends on how a decision backbone encodes candidate options prior to scalar logit projection. Contemporary single-pass System-1 decision models fall into three structural families:

\begin{enumerate}[leftmargin=1.5em]
    \item \textbf{Independent Causal Suffix Scoring (\texttt{Qwen3-0.6B-RLCD} \citep{jeff2026}):} The context state $s$ is prefilled once into a read-only causal KV prefix cache. Each candidate $o_i \in \mathcal{O}$ is scored as an independent suffix branch conditioned on the shared prefix cache, and $\mathbf{e}_i \in \mathbb{R}^d$ is extracted by mean-pooling final-layer hidden states over $o_i$'s token span. Because candidate branches never attend to one another, injecting $m$ clones leaves non-cloned embeddings $\mathbf{e}_i$ and pre-readout logits $z_i$ strictly unchanged ($\partial z_i / \partial m = 0$), restricting failure strictly to \textbf{Level-1 readout crowding}.
    \item \textbf{Joint Bidirectional Encoders (\texttt{Laya-421M} / ModernBERT \citep{laya2026,modernbert2024}):} The state $s$, instructions, and all $K$ candidates are concatenated into a single sequence (\texttt{[CLS] <type> instructions [SEP] [MASK] o\_1 ... [MASK] o\_K [SEP] state [SEP]}) and encoded across 24 bidirectional self-attention layers, with $\mathbf{e}_i \in \mathbb{R}^{1024}$ gathered at each option's \texttt{[MASK]} marker. Because every option token attends bidirectionally to all other candidates across all layers, duplicating a distractor alters intermediate hidden states for every option in $\mathcal{O}$, inducing \textbf{Level-2 cross-option attention entanglement} prior to readout.
    \item \textbf{Cross-Option Matrix Stacks (\texttt{Jeva-4B} \citep{jeva2026,ndmblock2026}):} Context and option texts are encoded separately, but their slot representations subsequently pass through a multi-layer cross-candidate Matrix Decision Stack (depth 4--8, $d = 768$) performing self-attention across candidate slots before linear scoring, creating multi-stage \textbf{Level-2 matrix-stack entanglement}.
\end{enumerate}

\subsection{Mechanistic Margin Inversion Bound}
\label{sec:arch_bound}

In Level-2 entangled backbones, cross-option self-attention among $m+1$ duplicate distractor spans inflates the pre-readout clone logit $z_{\text{clone}}$ relative to the target logit $z^*$. We derive the exact condition required for the Gram readout discount $-\log c_i$ \eqref{eq:gram_head} to rescue top-1 accuracy.

\begin{proposition}[Readout Margin Inversion Bound]
    \label{prop:margin_bound}
    Assume $T = 1$, target effective crowd size $c^* = 1$ ($\tilde{z}^* = z^*$), and cloned distractor effective crowd size $c_{\text{clone}} \le m + 1$ (since $S_{ij} \in [0, 1]$). Then $\tilde{z}^* > \tilde{z}_{\text{clone}} = z_{\text{clone}} - \log c_{\text{clone}}$ holds if and only if the pre-readout logit margin deficit $\Delta z \triangleq z_{\text{clone}} - z^*$ satisfies:
    \begin{equation}
        \Delta z \triangleq z_{\text{clone}} - z^* < \log c_{\text{clone}} \le \log(m + 1).
        \label{eq:margin_bound}
    \end{equation}
\end{proposition}

Proposition~\ref{prop:margin_bound} explains why readout-only discounting succeeds on causal suffix scorers yet fails on frozen joint encoders. On \texttt{Qwen3-0.6B-RLCD}, independent suffix scoring preserves $z_{\text{clone}} = z_K^{(0)} < z^*$ on clean-correct queries ($\Delta z < 0 < \log(m+1)$), enabling the zero-shot Gram head to recover +24.0\% accuracy at $m=3$ and +17.0\% at $m=5$. Conversely, in frozen \texttt{Laya-421M} at $m=5$ ($\log 6 \approx 1.79\text{ nats}$), bidirectional self-attention across the $m+1=6$ duplicate spans inverts 32/100 clean-correct target rankings before readout (mean $\Delta z = 1.19\text{ nats}$, range $[0.03, 2.91]\text{ nats}$, with 6/32 exceeding $\log 6$), and simultaneously entangles the target representation itself so $c^* > 1$: on those 32 flips, mean target crowd size reaches $c^* = 6.60$ versus $c_{\text{clone}} = 5.21$ at $\gamma=4.0, \tau_{\text{sim}}=0.8$ (7.89 vs.\ 6.97 at $\gamma=1.0, \tau_{\text{sim}}=0.3$), making the net readout adjustment $\log(c_{\text{clone}}/c^*)$ negative ($-0.324$ and $-0.134\text{ nats}$ on average) and rescuing 0/32 queries. Because internal bidirectional attention corrupts both pre-readout margins and target isolation, no readout-only head can restore top-1 accuracy without internal attention decoupling.

\subsection{Block-Diagonal Parallel Masking \& Question-Conditioned OSA}
\label{sec:arch_decouple}

\begin{figure}[t]
\centering
\resizebox{\textwidth}{!}{%
\begin{tikzpicture}[
    node distance=5mm and 7mm,
    font=\small,
    box/.style={draw, rounded corners=3pt, align=center, inner sep=5pt, minimum height=9mm, fill=white},
    redbox/.style={box, draw=entanglestroke, fill=entanglefill},
    bluebox/.style={box, draw=cleanstroke, fill=cleanfill},
    greenbox/.style={box, draw=decouplestroke, fill=decouplefill},
    arrow/.style={-{Stealth[length=2.2mm]}, thick}
]

\node[box, minimum width=6.1cm, fill=gray!6, draw=gray!60] (l_in) {\textbf{Input Sequence:} $[s;\; o_1;\; o_2;\; \dots;\; o_{K+m}]$};
\node[redbox, below=6mm of l_in, minimum width=6.1cm] (l_attn) {\textbf{Full Bidirectional Self-Attention}\\Unmasked option-to-option edges ($M_{ab}=1$)};
\node[redbox, below=6mm of l_attn, minimum width=6.1cm] (l_logit) {\textbf{Level-2 Entangled Logits \& States}\\32/100 flips; target $c^*=6.60 > c_{\text{clone}}$};
\node[redbox, below=6mm of l_logit, minimum width=6.1cm] (l_sm) {\textbf{Unweighted Softmax Readout}\\Level-1 $\mathcal{O}(1/m)$ dilution (56.0\% acc at $m=5$)};

\draw[arrow] (l_in) -- (l_attn);
\draw[arrow] (l_attn) -- (l_logit);
\draw[arrow] (l_logit) -- (l_sm);

\node[above=2mm of l_in, font=\bfseries\small, text=entanglestroke] {(a) Baseline: Joint Attention Entanglement};

\node[box, right=12mm of l_in, minimum width=8.2cm, text width=7.9cm, fill=gray!6, draw=gray!60] (r_in) {\textbf{Pre-Attention Token Pooling} $\mathbf{u}_i = \frac{1}{|T(o_i)|}\sum_{t}\mathbf{E}_{\text{tok}}[t]$\\Centered $s_{\max} = \max_{i\ne j}\hat{\mathbf{u}}_i^\top\hat{\mathbf{u}}_j$ (\texttt{CLINC150}: $m=0$: $0.160$; $m=5$: $0.756$)};

\node[diamond, draw=gray!70!black, fill=yellow!10, aspect=2.4, inner sep=2.5pt, align=center, below=6mm of r_in, font=\footnotesize\bfseries] (r_gate) {$s_{\max} \ge \tau_{\text{gate}}$?\\($\tau_{\text{gate}}=0.50$)};

\node[bluebox, below=10mm of r_gate, xshift=-2.45cm, minimum width=3.4cm, text width=3.2cm, font=\footnotesize] (r_seq) {\textbf{Sequential Path}\\($s_{\max} < 0.50$)\\Base Encoder + Softmax\\88.0\% clean ($m=0$)};

\node[greenbox, below=10mm of r_gate, xshift=2.45cm, minimum width=4.0cm, text width=3.8cm, font=\footnotesize] (r_par) {\textbf{Block-Diagonal Mask + LoRA}\\($s_{\max} \ge 0.50$)\\$M_{ab}=0$ cross-option; layers 20--23};

\node[greenbox, below=5mm of r_par, minimum width=4.0cm, text width=3.8cm, font=\footnotesize] (r_osa) {\textbf{Question-Conditioned OSA}\\Global-mean centered $\tilde{\mathbf{e}}_i = \mathbf{e}'_i - \boldsymbol{\mu}_{\text{global}}$};

\node[greenbox, below=5mm of r_osa, minimum width=4.0cm, text width=3.8cm, font=\footnotesize] (r_gram) {\textbf{Gram-Discounted Head}\\$\tilde{z}_i = z_i/T - \log c_i$ (74.0\% at $m=5$)};

\draw[arrow] (r_in) -- (r_gate);
\draw[arrow] (r_gate.west) -| node[above left, pos=0.25, font=\scriptsize] {No (100\% $m{=}0$)} (r_seq.north);
\draw[arrow] (r_gate.east) -| node[above right, pos=0.25, font=\scriptsize] {Yes (100\% $m{=}5$)} (r_par.north);
\draw[arrow] (r_par) -- (r_osa);
\draw[arrow] (r_osa) -- (r_gram);

\node[above=2mm of r_in, font=\bfseries\small, text=decouplestroke] {(b) Proposed: Disentangled Similarity-Gated Routing};

\end{tikzpicture}
}
\caption{Architectural schematic contrasting (a) joint bidirectional self-attention entanglement against (b) pre-attention Disentangled Similarity-Gated Routing ($\tau_{\text{gate}} = 0.50$), which routes clean queries to the fast sequential readout and duplicate-heavy queries through Block-Diagonal Parallel-Mask LoRA, Question-Conditioned OSA, and the global-centered \texttt{GramDiscountedChoiceHead}.}
\label{fig:architecture}
\end{figure}
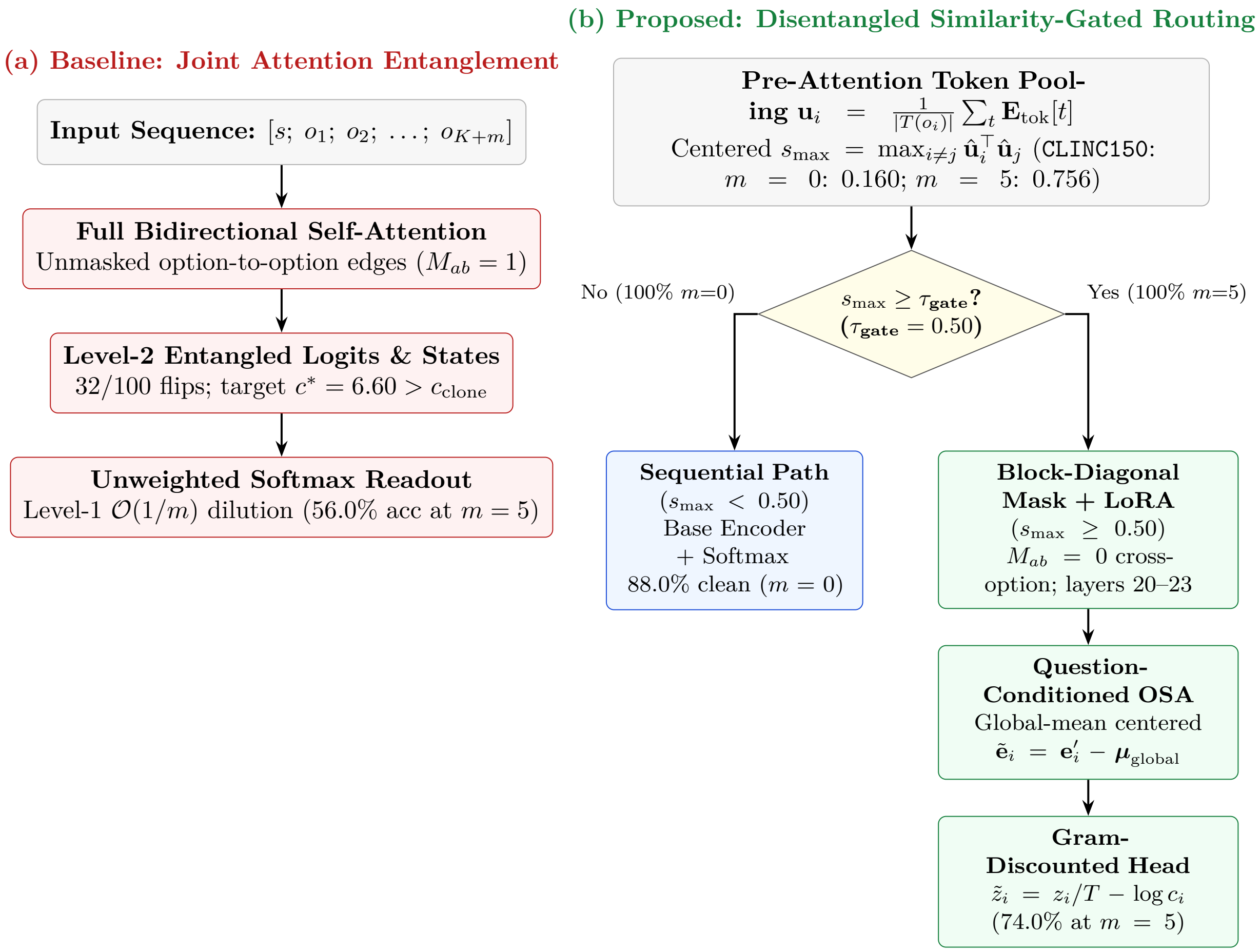

To eliminate Level-2 attention entanglement within a 4 GB VRAM budget, we combine block-diagonal parallel option masking with parameter-efficient adaptation (\textbf{Figure~\ref{fig:architecture}}). Let $\mathcal{I}_s = \{a : \text{opt}(a) = 0\}$ index shared context and instruction tokens, and let $\mathcal{I}_{o_i} = \{a : \text{opt}(a) = i\}$ index candidate span $o_i$ ($i \in \{1, \dots, K\}$) in $[s; o_1; \dots; o_K]$. Following parallel option masking \citep{laya2026}, we align positional indices so every option span $\mathcal{I}_{o_i}$ begins at identical position offset $p_{\text{start}}$ after the instruction \texttt{[SEP]}, and enforce a 2D block-diagonal binary attention mask $M \in \{0, 1\}^{L \times L}$:
\begin{equation}
    M_{ab} = \begin{cases} 0, & \text{if } \text{opt}(a) > 0 \;\wedge\; \text{opt}(b) > 0 \;\wedge\; \text{opt}(a) \ne \text{opt}(b), \\ 1, & \text{otherwise}. \end{cases}
    \label{eq:parallel_mask}
\end{equation}
Under Eq.~\eqref{eq:parallel_mask}, option tokens $a \in \mathcal{I}_{o_i}$ attend strictly to $\mathcal{I}_s \cup \mathcal{I}_{o_i}$ and are masked out ($M_{ab} = 0$) from all competing spans $\mathcal{I}_{o_j}$ ($j \ne i$). Because each option is processed identically to $m=0$, pre-readout representations become structurally invariant to distractor injection.

Because encoders pretrained with joint attention lose comparative calibration when cross-option edges are masked, we freeze all 421M backbone parameters and attach low-rank adapters (LoRA \citep{hu2021lora}) of rank $r_{\text{LoRA}} = 8$ ($\alpha_{\text{LoRA}} = 16.0$) to the fused $\mathbf{W}_{qkv} \in \mathbb{R}^{3d \times d}$ projections in the top 4 ModernBERT layers (\texttt{layers 20--23}; 131,072 parameters). At the readout boundary, we introduce a \textbf{Question-Conditioned Orthogonal Set Adapter (OSA)} (199,809 parameters) that conditions option embeddings $\mathbf{e}_i \in \mathbb{R}^d$ ($d = 1024$) on the \texttt{[CLS]} context representation $\mathbf{h}_q \in \mathbb{R}^d$ with bottleneck rank $r_{\text{OSA}} = 64$:
\begin{equation}
    \mathbf{h}_i = \text{GELU}\!\left(\mathbf{W}_1\,\text{LN}(\mathbf{e}_i) + \mathbf{W}_q \mathbf{h}_q\right), \qquad \mathbf{e}'_i = \mathbf{e}_i + \alpha\,\mathbf{W}_2 \mathbf{h}_i, \qquad z_i = z_{\text{base}, i} + \mathbf{w}_{\text{logit}}^\top \mathbf{h}_i,
    \label{eq:osa}
\end{equation}
where $\mathbf{W}_1, \mathbf{W}_q \in \mathbb{R}^{r_{\text{OSA}} \times d}$, $\mathbf{W}_2 \in \mathbb{R}^{d \times r_{\text{OSA}}}$, $\mathbf{w}_{\text{logit}} \in \mathbb{R}^{r_{\text{OSA}}}$ (zero-initialized), and $\alpha \in \mathbb{R}$ (initialized to 0.1), followed by Global Corpus Mean Centering \eqref{eq:global_mean} and the \texttt{GramDiscountedChoiceHead} \eqref{eq:gram_head}. The combined architecture (\texttt{Variant B}) requires 330,881 trainable parameters ($<0.08\%$ of backbone weights, 1,896.4 MB peak VRAM).

We train $\theta = \{\text{LoRA}, \text{OSA}\}$ on the disjoint 600-query \texttt{CLINC150} split (optimization details in Appendix~\ref{app:impl}) under the composite objective:
\begin{equation}
    \mathcal{L}_{\text{total}}(\theta) = \mathcal{L}_{\text{CE}} + \lambda_{\text{Brier}}\mathcal{L}_{\text{Brier}} + \lambda_{\text{Ortho}}\mathcal{L}_{\text{Ortho}} + \lambda_{\text{Inv}}\mathcal{L}_{\text{Inv}} + \lambda_{\text{Anchor}}\mathcal{L}_{\text{Anchor}},
    \label{eq:loss}
\end{equation}
combining target cross-entropy $\mathcal{L}_{\text{CE}} = -\log P_\theta(o^* \mid s, \mathcal{O}^{(m)})$, Brier calibration loss $\mathcal{L}_{\text{Brier}} = \sum_{i} (P_\theta(o_i) - y_i)^2$ \citep{brier1950}, distinct-pair orthogonality loss $\mathcal{L}_{\text{Ortho}}$, invariance penalty $\mathcal{L}_{\text{Inv}} = (1 - P_\theta(o^*))^2$, and clean-teacher KL anchor $\mathcal{L}_{\text{Anchor}}$:
\[
\mathcal{L}_{\text{Ortho}} = \frac{1}{|\mathcal{P}_{\text{dist}}|}\sum_{(i,j)\in\mathcal{P}_{\text{dist}}} (\hat{\mathbf{e}}_i'^\top \hat{\mathbf{e}}_j')^2, \qquad
\mathcal{L}_{\text{Anchor}} = D_{\text{KL}}\!\left(P_{\text{seq}}^{(0)}(\cdot \mid s, \mathcal{O}^{(0)}) \,\|\, P_\theta(\cdot \mid s, \mathcal{O}^{(0)})\right),
\]
with weights $\lambda_{\text{Brier}} = 0.10, \lambda_{\text{Ortho}} = 0.15, \lambda_{\text{Inv}} = 0.10, \lambda_{\text{Anchor}} = 0.50$. \textbf{Figure~\ref{fig:architecture}} depicts the architectural schematic contrasting joint self-attention against Block-Diagonal Parallel Masking, Question-Conditioned OSA, and Disentangled Gated Routing.

\subsection{Disentangled Similarity-Gated Inference Routing}
\label{sec:arch_gating}

Unconditional parallel masking eliminates cross-option comparison even on clean ($m=0$) queries where direct candidate contrast aids discrimination (88.0\% sequential vs.\ 78.0\% unconditional Variant B). We therefore define a runtime routing policy $\text{Policy}(s, \mathcal{O})$ using maximum off-diagonal candidate similarity $s_{\max} = \max_{i \ne j} \hat{\mathbf{u}}_i^\top \hat{\mathbf{u}}_j$:
\begin{equation}
    \text{Policy}(s, \mathcal{O}) = \begin{cases} \text{Parallel-Mask LoRA + OSA + Gram Head (Variant B)}, & \text{if } s_{\max} \ge \tau_{\text{gate}}, \\ \text{Standard Sequential Encoder Readout}, & \text{if } s_{\max} < \tau_{\text{gate}}. \end{cases}
    \label{eq:policy}
\end{equation}

A diagnostic sweep shows that computing $s_{\max}$ from post-attention sequential hidden states $\mathbf{e}_i$ fails due to bidirectional attention anisotropy: clean $m=0$ queries exhibit mean $s_{\max} = 0.9856$ on the Stress Test and 0.9609 on \texttt{Banking77}, forcing the 95th-percentile threshold to $\tau^*_{\text{gate}} = 0.9999$ (which fails to trigger at $m=5$, dropping accuracy to 58.0\%) while $\tau_{\text{gate}} = 0.50$ misroutes 100.0\% of clean queries (88.0\% $\to$ 78.0\%). Extracting $\mathbf{u}_i$ via mean pooling of input token embeddings $\mathbf{E}_{\text{tok}}$ \textbf{prior} to cross-option self-attention---centered by the clean $m=0$ token-embedding mean $\bar{\mathbf{u}}_{\text{clean}}$:
\begin{equation}
    \mathbf{u}_i = \frac{1}{|T(o_i)|}\sum_{t \in T(o_i)} \mathbf{E}_{\text{tok}}[t], \qquad \hat{\mathbf{u}}_i = \frac{\mathbf{u}_i - \bar{\mathbf{u}}_{\text{clean}}}{\|\mathbf{u}_i - \bar{\mathbf{u}}_{\text{clean}}\|_2}, \qquad s_{\max} = \max_{i \ne j} \hat{\mathbf{u}}_i^\top \hat{\mathbf{u}}_j
    \label{eq:gate_sim}
\end{equation}
decouples gating from sequence entanglement, yielding clean bimodal separation: clean $m=0$ queries lie in $[0.048, 0.259]$ (mean 0.160, $100\% < 0.50$ on Stress Test; mean 0.165, $99\% < 0.50$ on \texttt{Banking77}), whereas $m=5$ paraphrase clones lie in $[0.683, 0.868]$ (mean 0.756, $100\% \ge 0.50$) on the Stress Test and $[0.536, 0.948]$ (mean 0.843, $100\% \ge 0.50$) on \texttt{Banking77}, enabling $\tau_{\text{gate}} = 0.50$ to preserve 100\% of clean baseline accuracy while routing 100\% of clone attacks to Variant B.

The Gram readout inside \texttt{QuestionConditionedOSA} uses the module defaults throughout adapter training and evaluation ($\tau_{\text{sim}} = 0.50$, $\gamma = 2.0$). The milestone-2 grid optimum ($\gamma = 1.0$, $\tau_{\text{sim}} = 0.30$) applies only to the readout-only head on \texttt{Laya-421M} and was not used to train or evaluate Variants A or B.

\section{Evaluation Protocol \& Benchmark Design}
\label{sec:eval}

\subsection{Backbone Selection \& Native Prompt Standardization}
\label{sec:eval_backbones}

We evaluate one backbone from each family in Section~\ref{sec:arch_families}. \texttt{Laya-421M} (\texttt{convaiinnovations/laya}) is a 24-layer ModernBERT bidirectional joint encoder with $d=1024$ \citep{laya2026,modernbert2024}. \texttt{Qwen3-0.6B-RLCD} (\texttt{anthonym21/qwen3-0.6b-rlcd-decision}) is a 0.6B causal decoder that shares a prefix KV-cache and scores each option suffix independently \citep{jeff2026}. \texttt{Jeva-4B} (\texttt{LIQIIIII/Jeva-4B}) is a 4B decoupled text encoder with a cross-candidate Matrix Decision Stack of depth 4 and $d=768$ \citep{jeva2026,ndmblock2026}.

Each model is run strictly in its native prompt schema (exact templates in Appendix~\ref{app:data}), including \texttt{Jeva-4B}'s structured key-description decision rubric. This keeps option-splitting failures attributable to architecture rather than prompt mismatch.

\subsection{Benchmark Suites \& Leakage Governance}
\label{sec:eval_suites}

The protocol uses 1,300 paired evaluation queries in three suites, plus a separate 600-query adapter training split (Table~\ref{tab:benchmarks}).

\begin{table}[t]
\centering
\caption{Benchmark summary.}
\label{tab:benchmarks}
\small
\begin{tabularx}{\textwidth}{L{2.5cm} Y L{1.9cm} L{1.2cm} Y Y}
\toprule
Suite & Domain \& Source & Eval Queries & Clean $K$ & Perturbation Regimes & Leakage Control \\
\midrule
Controlled Stress Test & \texttt{CLINC150} \citep{larson2019clinc}, 8 support / incident / triage domains & 900 & 4 & Paraphrase cloning, sub-splitting ($m \le 5$) & Fixed; disjoint from training \\
Banking77 Routing & \texttt{Banking77} \citep{casanueva2020banking77} & 200 (100 pairs) & 4 & Paraphrase cloning, $m \in \{0, 5\}$ & Independent dataset \\
OpenDecisions & 11 task families \citep{jeff2026} & 200 (100 pairs) & 3--6 & Paraphrase cloning, $m \in \{0, 5\}$ & Independent dataset \\
Adapter Training Split & \texttt{CLINC150} Travel, Home, Kitchen/Dining & 600 (480 train / 120 val) & 3--8 & $m \in \{0,1,2,3\}$ & 0\% lexical, scenario, or intent overlap with the stress test \\
\bottomrule
\end{tabularx}
\end{table}

The Controlled Stress Test starts from 100 base scenarios with one target and three distinct distractors (benchmark schemas and disjointness details in Appendix~\ref{app:data}). Each base scenario is evaluated in 9 paired slices: a clean control ($m=0$), four paraphrase-clone slices ($m \in \{1,2,3,5\}$, $K \in \{5,6,7,9\}$), and four sub-split slices ($m \in \{1,2,3,5\}$, $K \in \{4,5,6,8\}$). A separate 100-query positive-target cloning check duplicates the ground-truth option.

\texttt{Banking77} tests high-competence intent routing across all 77 banking classes: clean accuracy is 93.0--94.0\% for \texttt{Laya-421M} and 75.0\% for \texttt{Qwen3-0.6B-RLCD}. \texttt{OpenDecisions} contains multi-paragraph tasks including LLM routing, fairness, guardrail classification, and multilingual decisions.

The training split was filtered by domain and keyword so that zero scenarios or intents overlap with the stress test. Its clean ($m=0$) queries also supply the global background mean $\boldsymbol{\mu}_{\text{global}} \in \mathbb{R}^{1024}$ for Section~\ref{sec:prob_centering} centering. A deterministic alignment audit (Appendix~\ref{app:data}) confirms exact one-to-one base-ID pairing between clean and perturbed slices.

\subsection{Perturbation Taxonomy}
\label{sec:eval_perturb}

\textbf{Distractor paraphrase cloning.} We append $m \in \{0,1,2,3,5\}$ paraphrases of one non-target distractor $o_{\text{dist}} \neq o^*$, so $K \to K+m$. The context $s$ and target $o^*$ are unchanged. Under the clone-invariance axioms of Section~\ref{sec:prob_axioms}, the target probability should not change.

\textbf{Taxonomic sub-splitting.} We replace $o_{\text{dist}}$ with $m+1$ mutually exclusive, fine-grained children. These options are legitimately distinct, so a head that discounts them as clones is over-pruning. This slice measures over-pruning under continuous Gram discounting on hierarchically distinct alternatives.

\textbf{Positive-target cloning.} We replicate $o^*$ into an equivalence class $\mathcal{C}^*$ of paraphrased copies, up to $\lvert\mathcal{C}^*\rvert = 6$ ($K=9$), with distractors fixed. This checks the symmetry in Definition~\ref{def:clone_inv}: a clone-invariant head should conserve class mass $\sum_{c \in \mathcal{C}^*} P(c)$ rather than inflate it through additional Softmax terms.

\subsection{Baselines, Metrics \& Statistical Protocol}
\label{sec:eval_protocol}

We compare six configurations: (1) standard Softmax; (2) scalar temperature scaling ($T^*=1.11$, fit by NLL on clean validation logits; \citealp{guo2017calibration}); (3) zero-shot \texttt{GramDiscountedChoiceHead} with global mean centering (module default $\gamma=2.0, \tau_{\text{sim}}=0.50$ and milestone-2 grid optimum $\gamma=1.0, \tau_{\text{sim}}=0.30$; per-benchmark settings in Appendix~\ref{app:impl}: \texttt{Laya-421M} stress test $\gamma=4.0, \tau_{\text{sim}}=0.8$, \texttt{Qwen3-0.6B-RLCD} on \texttt{CLINC150} and \texttt{OpenDecisions} $\gamma=4.0, \tau_{\text{sim}}=0.5$, and \texttt{Banking77} / \texttt{Jeva-4B} $\gamma=2.0, \tau_{\text{sim}}=0.5$); (4) readout-only OSA (Variant A, 199,809 parameters); (5) parallel-mask LoRA + OSA (Variant B, 330,881 parameters); and (6) Variant B with similarity-gated routing at $\tau_{\text{gate}}=0.50$.

We report the following metrics:
\begin{itemize}[leftmargin=1.5em]
    \item \textbf{Top-1 accuracy:} $\frac{1}{N}\sum_n \mathbb{I}[\operatorname{argmax}_i P_n(o_i) = i^*_n]$.
    \item \textbf{Cluster top-1 accuracy:} under positive-target cloning, the argmax lies in $\mathcal{C}^*_n$.
    \item \textbf{Decision flip rate (DFR):} $\frac{1}{N}\sum_n \mathbb{I}[\hat{y}_n^{(m)} \neq \hat{y}_n^{(0)}]$, the fraction of paired queries whose top-1 choice changes relative to the unmodified baseline's clean prediction. For unadapted baselines at $m=0$, DFR against their own clean predictions is identically 0.0\%; for adapted models (\texttt{Variant B}), $m=0$ DFR measures disagreement against the unmodified Softmax baseline's clean prediction ($\mathbb{I}[\hat{y}_{n,\text{adapted}}^{(0)} \neq \hat{y}_{n,\text{base}}^{(0)}]$), accounting for non-zero clean flip rates when parallel masking is always active.
    \item \textbf{15-bin ECE} \citep{guo2017calibration}: $\sum_{b=1}^{15} \frac{|B_b|}{N}\,|\text{acc}(B_b) - \text{conf}(B_b)|$.
    \item \textbf{Multi-class Brier score} \citep{brier1950}: $\frac{1}{N}\sum_n \sum_i (P_n(o_i) - y_{n,i})^2$.
    \item \textbf{Mean target probability} $\bar{P}(o^*)$ and \textbf{distractor cluster mass} $\sum_{c \in \mathcal{C}_{\text{dist}}} P(c)$.
\end{itemize}

Confidence intervals are 95\% percentile bootstrap intervals over paired resamples ($B=1{,}000$, seed 42). Adapter training uses three seeds (42, 123, 777). Thirteen automated unit tests (Appendix~\ref{app:tests}) check the probability and invariance properties. All runs fit within a 4 GB VRAM budget (NVIDIA RTX 3050 Laptop GPU).

\section{Empirical Results \& Diagnostic Analysis (RQ1--RQ5)}
\label{sec:results}

\subsection{RQ1: Controlled Stress Test \& Level-2 Gated Recovery}
\label{sec:results_rq1}

Option splitting cuts \texttt{Laya-421M} top-1 accuracy from 88.0\% to 56.0\% at $m=5$ (Table~\ref{tab:stress}), a 40.0\% decision flip rate, and readout-only fixes do not move it. Each slice holds 100 paired queries.

\begin{table}[t]
\centering
\caption{Controlled stress test on \texttt{Laya-421M}, paraphrase cloning (accuracy \%, 100 queries per slice). Brackets give 95\% bootstrap CIs where logged. ``Tax.\ $m{=}5$'' is accuracy under taxonomic sub-splitting. The zero-shot Gram row uses $\gamma=4.0$, $\tau_{\text{sim}}=0.8$ (Section~\ref{sec:eval_protocol}).}
\label{tab:stress}
\resizebox{\textwidth}{!}{%
\begin{tabular}{lccccccc}
\toprule
Method & $m{=}0$ & $m{=}3$ & $m{=}5$ & DFR $m{=}5$ & $\bar P(o^*)$ $m{=}5$ & Distractor mass $m{=}5$ & Tax.\ $m{=}5$ \\
\midrule
Softmax & 88.0 & 70.0 & 56.0 [46.0, 66.0] & 40.0 & 0.4308 & 0.4852 & 77.0 \\
Temperature scaling ($T^*=1.11$) & 88.0 & 70.0 & 56.0 & 40.0 & 0.4141 & 0.4990 & 77.0 \\
Zero-shot Gram head & 88.0 & 70.0 & 56.0 & 40.0 & 0.4648 & 0.4396 & 77.0 \\
Variant A (readout-only OSA) & 88.0 & 70.0 & 56.0 & 40.0 & 0.5189 & 0.4053 & 77.0 \\
Variant B (LoRA + OSA, always on) & 78.0 & 80.0 & 74.0 [64.97, 82.0] & 22.0 & 0.7095 & 0.0457 & 74.0 \\
Variant B, similarity-gated & 88.0 & 80.0 & 74.0 [64.97, 82.0] & 22.0 & 0.7095 & 0.0457 & 85.0 \\
\bottomrule
\end{tabular}%
}
\end{table}

Temperature scaling leaves accuracy and DFR unchanged at every $m$, as Proposition~\ref{prop:temp} requires; it only rescales confidence. The zero-shot Gram head and Variant A also leave accuracy at 56.0\% (distractor mass falls from 0.4852 to 0.4396 and 0.4053). This is the margin inversion of Proposition~\ref{prop:margin_bound}: on the 32/100 clean-correct queries that invert at $m=5$, bidirectional attention yields a mean pre-readout deficit of $\Delta z = 1.19$ nats (with only 6/32 exceeding $\log 6 \approx 1.79$ nats) while simultaneously entangling the target representation ($c^* = 6.60 > c_{\text{clone}} = 5.21$, mean net readout adjustment $-0.324$ nats), preventing a post-hoc discount on frozen logits from restoring the target to the top.

Variant B changes the encoder pass and recovers 18.0 points at $m=5$ (56.0\% to 74.0\%), cuts DFR from 40.0\% to 22.0\%, and shrinks distractor mass $10.6\times$ (0.4852 to 0.0457). Used unconditionally it costs 10.0 points on clean queries (88.0\% to 78.0\%). The similarity gate removes that cost: all 100 clean queries route to the unmodified path (88.0\%, DFR 0.0\%), and all 400 cloned queries route to Variant B. Clean accuracy is therefore preserved by construction, not learned. The gate also does what hard routing should on sub-splits: only 38\% of $m=5$ sub-split queries are routed to Variant B, and accuracy is 85.0\% against 77.0\% for Softmax and 74.0\% for always-on Variant B. The 95\% bootstrap intervals for Softmax and Variant B at $m=5$ overlap slightly ([46.0, 66.0] vs.\ [64.97, 82.0]), so we report the +18.0-point recovery alongside consistent improvements across all $m \ge 3$ regimes rather than claiming statistical separation under a formal paired test. In addition, Variant B trades off low-multiplicity accuracy when active unconditionally: at $m=1$ and $m=2$ it scores 81.0\% and 80.0\% compared to Softmax at 88.0\% and 87.0\%, overtaking Softmax once duplicate crowding becomes severe ($m \ge 3$).

\subsection{RQ2: Cross-Architecture Generalization Across Real-World Benchmarks}
\label{sec:results_rq2}

The two intervention levels help different backbones, and neither helps a backbone that is wrong to begin with. Table~\ref{tab:cross_arch} reports accuracy at $m=0 \to m=5$, DFR at $m=5$, and distractor mass at $m=5$.

\begin{table}[t]
\centering
\caption{Cross-architecture results (100 queries per cell). Zero-shot Gram is the Level-1 head; Gated B is Variant B with similarity gating, trained on \texttt{CLINC150} only. \texttt{Jeva-4B} uses its native rubric prompt. Gram settings per row: \texttt{Qwen3-0.6B-RLCD} on \texttt{CLINC150} and \texttt{OpenDecisions} $\gamma=4.0$, $\tau_{\text{sim}}=0.5$; all other rows $\gamma=2.0$, $\tau_{\text{sim}}=0.5$ (Section~\ref{sec:eval_protocol}).}
\label{tab:cross_arch}
\small
\begin{tabular}{lllccc}
\toprule
Benchmark & Backbone & Method & Acc $m{=}0 \to m{=}5$ & DFR $m{=}5$ & Mass $m{=}5$ \\
\midrule
CLINC150 & Qwen3-0.6B-RLCD & Softmax & 96.0 $\to$ 57.0 & 43.0 & 0.4886 \\
CLINC150 & Qwen3-0.6B-RLCD & Zero-shot Gram & 96.0 $\to$ 74.0 & 26.0 & 0.4479 \\
Banking77 & Qwen3-0.6B-RLCD & Softmax & 75.0 $\to$ 43.0 & 47.0 & 0.5434 \\
Banking77 & Qwen3-0.6B-RLCD & Zero-shot Gram & 72.0 $\to$ 51.0 & 40.0 & 0.3101 \\
Banking77 & Laya-421M & Softmax & 93.0 $\to$ 69.0 & 27.0 & 0.4065 \\
Banking77 & Laya-421M & Zero-shot Gram & 93.0 $\to$ 69.0 & 27.0 & 0.3701 \\
Banking77 & Laya-421M & Gated B & 94.0 $\to$ 72.0 & 23.0 & 0.1243 \\
Banking77 & Jeva-4B & Softmax & 30.0 $\to$ 11.0 & 72.0 & 0.7071 \\
Banking77 & Jeva-4B & Zero-shot Gram & 30.0 $\to$ 12.0 & 71.0 & 0.6998 \\
OpenDecisions & Qwen3-0.6B-RLCD & Softmax & 27.0 $\to$ 17.0 & 44.0 & 0.5377 \\
OpenDecisions & Qwen3-0.6B-RLCD & Zero-shot Gram & 27.0 $\to$ 16.0 & 44.0 & 0.5161 \\
OpenDecisions & Laya-421M & Softmax & 41.0 $\to$ 27.0 & 42.0 & 0.5980 \\
OpenDecisions & Laya-421M & Always-on B & 39.0 $\to$ 31.0 & 46.0 & 0.4067 \\
OpenDecisions & Jeva-4B & Softmax & 32.0 $\to$ 13.0 & 62.0 & 0.6765 \\
OpenDecisions & Jeva-4B & Zero-shot Gram & 32.0 $\to$ 13.0 & 63.0 & 0.6812 \\
\bottomrule
\end{tabular}
\end{table}

\begin{figure}[t]
    \centering
    \includegraphics[width=\textwidth]{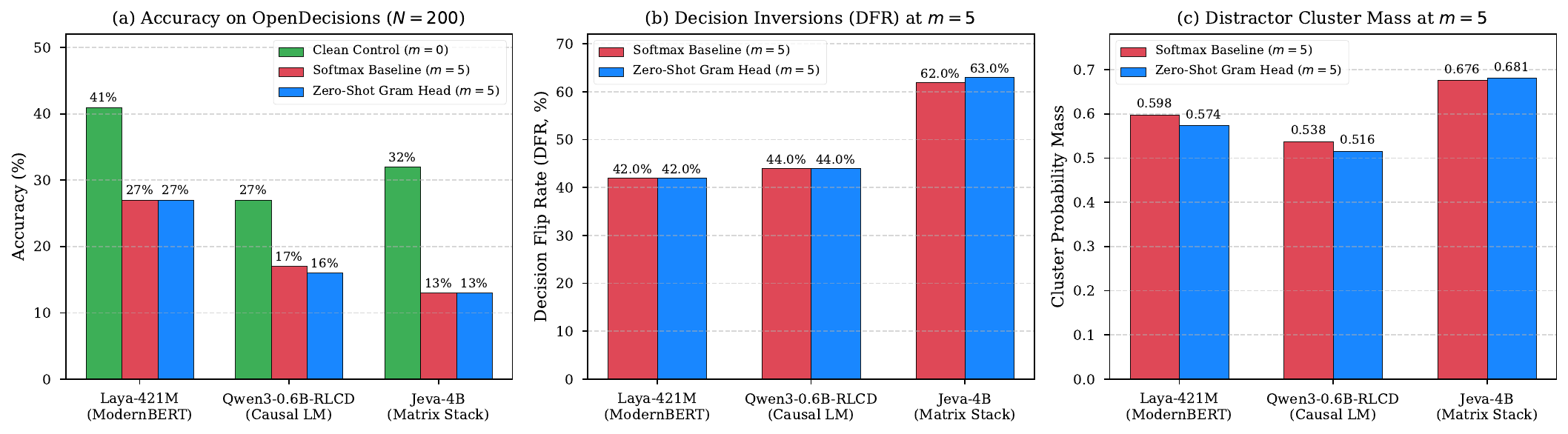}
    \caption{Cross-architecture comparison across \texttt{Laya-421M}, \texttt{Qwen3-0.6B-RLCD}, and \texttt{Jeva-4B} on \texttt{OpenDecisions} ($N=200$) under $m=0$ and $m=5$ option cloning.}
    \label{fig:cross_arch}
\end{figure}

\textbf{Suffix scorers.} On \texttt{Qwen3-0.6B-RLCD}, the Level-1 head alone recovers 24.0 points at $m=3$ on \texttt{CLINC150} (63.0\% to 87.0\%) and 17.0 points at $m=5$ (57.0\% to 74.0\%), while preserving high accuracy on \texttt{CLINC150} taxonomic sub-splits at $m=5$ (86.0\% [79.0, 92.0] vs.\ 85.0\% [77.0, 92.0] for Softmax, DFR 14.0\% vs.\ 15.0\%). On \texttt{Banking77}, it recovers 8.0 points (43.0\% to 51.0\%) and cuts distractor mass from 0.5434 to 0.3101, at a 3.0-point clean cost (75.0\% to 72.0\%). This matches the Section~\ref{sec:arch_families} prediction that independent suffix scoring suffers pure Level-1 crowding.

\textbf{Joint encoder.} On \texttt{Laya-421M}, the Level-1 head leaves \texttt{Banking77} accuracy at 69.0\% (distractor mass 0.4065 to 0.3701). Gated Variant B, trained on \texttt{CLINC150} and applied without adaptation, keeps 94.0\% clean accuracy and reaches 72.0\% at $m=5$ (DFR 23.0\%, mass 0.1243). That is a 3-point gain over Softmax, much smaller than the 18-point gain on \texttt{CLINC150}. Distractor mass fell $3.3\times$ (0.4065 to 0.1243), so most of the transfer shows up in the probabilities rather than in the argmax.

\textbf{Where neither level helps.} On \texttt{OpenDecisions} (\textbf{Figure~\ref{fig:cross_arch}}), clean accuracy is 41.0\% for \texttt{Laya-421M} and 27.0\% for \texttt{Qwen3-0.6B-RLCD}. Readout discounting only helps when the target already leads the distractors, and here it often does not, so the Level-1 head does not change accuracy (27.0\% vs.\ 27.0\% on Laya; 16.0\% vs.\ 17.0\% on Qwen). Always-on Variant B lifts Laya from 27.0\% to 31.0\% at $m=5$, but clean accuracy drops from 41.0\% to 39.0\% and clean DFR is 30.0\%, so this is not a clean win. \texttt{Jeva-4B} falls from 32.0\% to 13.0\% on \texttt{OpenDecisions} and from 30.0\% to 11.0\% on \texttt{Banking77} with its native prompt, and the Gram head changes neither (13.0\%, 12.0\%). Its cross-candidate stack mixes options before the readout, which is the Level-2 entanglement of Section~\ref{sec:arch_families}. Extending block-diagonal decoupling to \texttt{Jeva-4B}'s multi-layer Matrix Decision Stack remains open for future work.

\subsection{RQ3: Representation Geometry \& Centering Pathology Ablation}
\label{sec:results_rq3}

Centering determines whether the similarity threshold can separate clones from distinct options at all. Table~\ref{tab:centering} gives mean cosine similarities on 100 $m=5$ paraphrase-cloned queries.

\begin{table}[t]
\centering
\caption{Centering ablation on \texttt{Laya-421M} (\textbf{Figure~\ref{fig:reliability_ablation}b}), with $\tau_{\text{sim}}=0.50$. Downstream columns use the readout-only Gram head at $\gamma=2.0$, $\tau_{\text{sim}}=0.50$ from the milestone-2 ablation grid (per-question centering is omitted downstream because intra-clone similarity collapses below threshold).}
\label{tab:centering}
\small
\begin{tabular}{lcccccc}
\toprule
Centering & Clone cos & Distinct cos & Gap & Acc $m{=}0$ & Acc $m{=}5$ & Mass $m{=}5$ \\
\midrule
Raw (uncentered) & 0.984 & 0.796 & 0.188 & 88.0 & 56.0 & 0.4738 \\
Per-question mean & 0.193 & $-$0.287 & 0.480 & --- & --- & --- \\
Global corpus mean & 0.931 & 0.463 & 0.468 & 88.0 & 57.0 & 0.4645 \\
\bottomrule
\end{tabular}
\end{table}

Raw embeddings place distinct options at 0.796, above $\tau_{\text{sim}}=0.50$, so every pair looks like a clone and the discount fires on clean queries. Per-question centering has the largest gap (0.480) but puts clones at 0.193, below the threshold: subtracting the question's own option mean removes the shared component that makes clones similar \eqref{eq:sample_mean}, so clones fall below $\tau_{\text{sim}}$. Global corpus centering keeps clones at 0.931 and distinct options at 0.463, on the correct side of the threshold, though the 0.037 margin to 0.50 is thin. Downstream, the effect is small: accuracy at $m=5$ moves from 56.0\% to 57.0\% and distractor mass from 0.4738 to 0.4645. Centering fixes clone detection; it does not clear the Proposition~\ref{prop:margin_bound} barrier.

\begin{figure}[t]
    \centering
    \includegraphics[width=\textwidth]{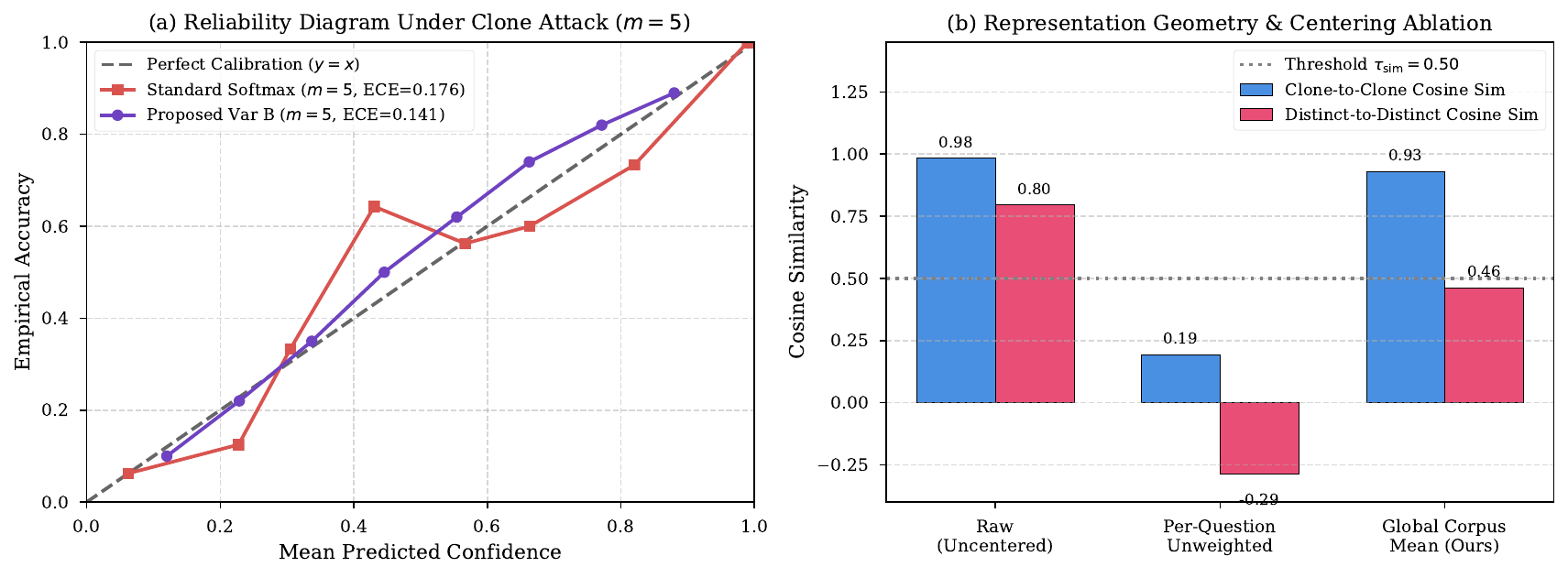}
    \caption{Calibration and representation geometry diagnostics on \texttt{Laya-421M}. Panel (a): 8-bin reliability diagram at $m=5$ comparing Standard Softmax against Variant B; Panel (b): Mean pairwise cosine similarity across centering schemes.}
    \label{fig:reliability_ablation}
\end{figure}

\subsection{RQ4: Cloning Symmetry \& Probability Calibration}
\label{sec:results_rq4}

Distractor cloning roughly doubles \texttt{Laya-421M} calibration error, and the variant that fixes accuracy gives up clean calibration. Softmax ECE rises from 0.0899 ($m=0$) to 0.1756 ($m=5$) and Brier from 0.1955 to 0.5679. Temperature scaling lowers clean ECE to 0.0569 but ends at 0.1577 ($m=5$, Brier 0.5700); the Gram head gives 0.0543 and 0.1611 (Brier 0.1880 and 0.5725). Always-on Variant B reaches ECE 0.1413 and Brier 0.3730 at $m=5$, a Brier reduction of 0.195 against an ECE reduction of 0.034. Its clean metrics are worse than Softmax (ECE 0.1062, Brier 0.3177), which is why gating matters for calibration as well as accuracy. \textbf{Figure~\ref{fig:reliability_ablation}a} plots reliability diagrams across 8 empirical confidence bins, while all ECE values reported in prose and Appendix~\ref{app:extended} follow the standard 15-bin estimator.

Duplicating the ground-truth option is the opposite attack: a clone-invariant head should hold the target class mass $\sum_{c \in \mathcal{C}^*} P(c)$ near its clean value. Table~\ref{tab:pos_target} reports the positive-target check at $m=5$ ($|\mathcal{C}^*|=6$, $K=9$, 100 queries), alongside clean $\bar P(o^*)$ from the stress-test control on the same base items.

\begin{table}[t]
\centering
\caption{Positive-target cloning at $m=5$. Clean $\bar P(o^*)$ is the $m=0$ control value; the zero-shot Gram head was not run on \texttt{Laya-421M} in this check.}
\label{tab:pos_target}
\small
\resizebox{\textwidth}{!}{%
\begin{tabular}{llcccc}
\toprule
Backbone & Method & Clean $\bar P(o^*)$ & Cluster top-1 (\%) & Target cluster mass & Distractor mass \\
\midrule
Laya-421M & Softmax & 0.8114 & 98.0 & 0.9470 & 0.0530 \\
Laya-421M & Variant B & 0.7530 & 75.0 & 0.7339 & 0.2661 \\
Qwen3-0.6B-RLCD & Softmax & 0.6289 & 100.0 & 0.7054 & 0.2946 \\
Qwen3-0.6B-RLCD & Zero-shot Gram & 0.6289 & 100.0 & 0.6799 & 0.3201 \\
\bottomrule
\end{tabular}%
}
\end{table}

Softmax inflates target mass above its clean value: 0.8114 to 0.9470 on Laya and 0.6289 to 0.7054 on Qwen. Variant B holds it at 0.7339 against a clean 0.7530, and the Gram head on Qwen reduces the inflation (0.6799) without removing it. Contrary to the concern that training exclusively on distractor clones might penalize duplicated targets, Variant B maintains 75.0\% cluster top-1 (within 3.0 points of its 78.0\% clean top-1) with only a 0.019 change in target cluster mass, redistributing residual probability mass compared to Softmax's sharp target concentration.

\subsection{RQ5: Multi-Seed Training Stability \& Hyperparameter Sensitivity}
\label{sec:results_rq5}

Always-on Variant B is stable across seeds but lower on clean queries than the single run in Table~\ref{tab:stress}. Over seeds 42, 123, and 777 (Table~\ref{tab:a3}), accuracy is $74.33 \pm 1.15\%$ at $m=0$ and $72.00 \pm 2.65\%$ at $m=5$, DFR is $20.67 \pm 2.52\%$, and distractor mass is $0.0602 \pm 0.0158$. The per-seed $m=5$ accuracies are 74.0\%, 73.0\%, and 69.0\%. The 78.0\% clean figure in Table~\ref{tab:stress} comes from a separate milestone-3 run and lies above all three sweep values, so we treat 74.33\% as the representative clean number for always-on Variant B; the gated system's 88.0\% clean accuracy does not depend on it. The seed spread (2.65 points at $m=5$) is a sizeable fraction of the 18-point gain.

Across the milestone-2 hyperparameter grid for the Level-1 head ($\gamma \in \{1, 2, 4\} \times \tau_{\text{sim}} \in \{0, 0.3, 0.5, 0.7\}$ with centering), $m=5$ accuracy remains between 55.0\% and 57.0\% and distractor mass between 0.4599 and 0.4670---confirming Proposition~\ref{prop:margin_bound}'s prediction that frozen joint encoders cannot be rescued by tuning post-readout penalties alone. For similarity gating, we evaluate the single theoretically motivated threshold $\tau_{\text{gate}}=0.50$ without post-hoc threshold tuning; an alternative threshold chosen from post-attention similarity calibration ($\tau^* \approx 0.9999$) routes only 53\% of $m=5$ queries to Variant B and drops accuracy to 58.0\%.

\section{Systems Guidance, Design Trade-offs \& Honest Limitations}
\label{sec:guidance}

\subsection{Practitioner Deployment Guide \& Systems Overhead}
\label{sec:guidance_deploy}

\texttt{GramDiscountedChoiceHead} adds zero trainable parameters, stores a single $d$-dimensional global mean vector (5,750 bytes at $d=1024$), and computes a $K \times K$ Gram matrix in $\mathcal{O}(K^2 d)$ operations, which is negligible at typical candidate cardinalities ($K \le 11$). Variant B adds 330,881 trainable parameters (131,072 LoRA and 199,809 OSA, $<0.08\%$ of \texttt{Laya-421M}) and trains in 274.5 s within 1,896.4 MB peak VRAM on a single 4 GB RTX 3050 Laptop GPU, leaving inference-latency microbenchmarking to future deployment profiling.

Always-on Variant B is not a drop-in replacement for Softmax. On \texttt{CLINC150}, Softmax keeps 88.0\% and 87.0\% at $m=1$ and $m=2$ against 81.0\% and 80.0\% for Variant B, and Variant B leads from $m \ge 3$ (74.0\% vs.\ 56.0\% at $m=5$). The pre-attention gate ($\tau_{\text{gate}}=0.50$) pools input token embeddings, an $\mathcal{O}(K)$ operation, and resolved this in our runs by sending all 100 clean queries to the sequential path (88.0\%) and all 400 paraphrase-cloned queries to Variant B. Table~\ref{tab:deployment} summarizes which intervention each backbone family supports.

\begin{table}[t]
\centering
\caption{Deployment decision matrix. Numbers are accuracy at $m=0 \to m=5$ from Tables~\ref{tab:stress}--\ref{tab:cross_arch} unless noted.}
\label{tab:deployment}
\small
\begin{tabularx}{\textwidth}{L{3.9cm} Y Y}
\toprule
Backbone family & Clean / low-redundancy options & High-duplicate / taxonomic-split options \\
\midrule
Independent causal suffix scorer (\texttt{Qwen3-0.6B-RLCD}) & Keep Softmax or add the Gram head; the head cost 0.0 points on \texttt{CLINC150} (96.0 $\to$ 96.0) and 3.0 on \texttt{Banking77} (75.0 $\to$ 72.0). & Gram head only: \texttt{CLINC150} 57.0 $\to$ 74.0, \texttt{Banking77} 43.0 $\to$ 51.0. Sub-splits unaffected (85.0 vs.\ 86.0 at $m=5$). \\
Joint bidirectional encoder (\texttt{Laya-421M}) & Route to the sequential readout (88.0\% \texttt{CLINC150}, 94.0\% \texttt{Banking77}). Always-on Variant B costs 10.0 points on \texttt{CLINC150}. & Gated Variant B: \texttt{CLINC150} 56.0 $\to$ 74.0, \texttt{Banking77} 69.0 $\to$ 72.0; sub-split $m=5$ 85.0 vs.\ 77.0. The Gram head alone gives 56.0 $\to$ 56.0. \\
Cross-option matrix stack (\texttt{Jeva-4B}) & Base accuracy is 30.0--32.0\% on \texttt{Banking77} and \texttt{OpenDecisions}; fix competence first. & No intervention validated: Softmax 32.0 $\to$ 13.0 and Gram 32.0 $\to$ 13.0 on \texttt{OpenDecisions}. Level 2 untested. \\
\bottomrule
\end{tabularx}
\end{table}

\subsection{Continuous Gram Discounting vs.\ Discrete Hard Pruning}
\label{sec:guidance_pruning}

Hard deduplication drops an option when $\max_{i \ne j} \hat{\mathbf{e}}_i^\top \hat{\mathbf{e}}_j \ge \tau$. Gram discounting instead sets $c_i = \sum_{j=1}^K S_{ij}^\gamma$ and $\tilde{z}_i = z_i/T - \log c_i$. Because our study isolates continuous discounting mechanisms rather than heuristic text deduplication, this comparison is structural rather than an empirical baseline.

Continuous discounting has three practical differences. It is differentiable, so the adapters in Variants A and B train through it. It keeps the caller's option count $K$ and the equivalence-class accounting of Definition~\ref{def:clone_inv}; in the positive-target check Variant B holds target cluster mass at 0.7339 against 0.7530 clean (Table~\ref{tab:pos_target}). And it grades partial similarity rather than deleting a candidate, which matters for sub-splits. A hard rule with a threshold below the sub-split similarity range would remove valid fine-grained options; our gated system scores 85.0\% on sub-split $m=5$, but the zero-shot Gram head does not (77.0\%), so that result comes from the routing, not the discount alone.

\subsection{Lexical vs.\ Semantic Gating Trade-off}
\label{sec:guidance_gating}

Pooling input token embeddings before cross-option attention separates clones from clean queries where post-attention states cannot. Post-attention similarities saturate (the calibrated threshold was $\tau^* \approx 0.9999$ and failed to trigger), while pre-attention scores split cleanly at $\tau_{\text{gate}}=0.50$: 100\% of clean stress-test queries and 99\% of clean \texttt{Banking77} queries fall below it. The cost is that the score is lexical. A paraphrase that shares no subword tokens with its source would yield a low $\hat{\mathbf{u}}_i^\top \hat{\mathbf{u}}_j$ and bypass the gate. Paraphrases sharing zero subword tokens remain an untested boundary case; pooling early contextual layers (layers 1--4) under the parallel attention mask offers a direct semantic upgrade without cross-option entanglement.

\subsection{Scope \& Honest Limitations}
\label{sec:guidance_limits}

\begin{enumerate}[leftmargin=1.5em]
    \item \textbf{Exact vs.\ approximate similarity.} Theorem~\ref{thm:clone_inv} assumes exact vector equality ($\cos = 1.0$), whereas real paraphrases exhibit mean centered cosine 0.931 (Table~\ref{tab:centering}), leaving moderate cross-seed variance ($72.00 \pm 2.65\%$ at $m=5$) and slightly overlapping 95\% bootstrap CIs against Softmax ([64.97, 82.0] vs.\ [46.0, 66.0]).
    \item \textbf{Base-competence floor.} Readout discounting and parallel adapters regulate candidate competition; when clean task competence is low (\texttt{OpenDecisions}, 27.0--41.0\% clean), reducing option crowding cannot recover queries where the backbone misidentifies the intent.
    \item \textbf{Static corpus mean and domain shift.} $\boldsymbol{\mu}_{\text{global}}$ is estimated once on \texttt{CLINC150} training options. Under cross-domain transfer to \texttt{Banking77}, gated Variant B improves $m=5$ accuracy by 3.0 points versus 18.0 in-domain (while still compressing distractor mass $3.3\times$); isolating how much of this accuracy gap stems from $\boldsymbol{\mu}_{\text{global}}$ drift versus domain-specific LoRA features remains open.
    \item \textbf{Clean calibration trade-off and fixed gate threshold.} Always-on Variant B lowers clean accuracy (78.0\% vs.\ 88.0\%) and clean calibration (ECE 0.1062 vs.\ 0.0899), making gating necessary; we evaluate $\tau_{\text{gate}}=0.50$ as a single fixed operating point rather than tuning thresholds across validation domains.
    \item \textbf{Untested matrix-stack decoupling on \texttt{Jeva-4B}.} Zero-shot evaluation confirms Level-2 vulnerability in \texttt{Jeva-4B}'s Matrix Decision Stack (32.0\% $\to$ 13.0\% on \texttt{OpenDecisions}); extending block-diagonal attention masking across multi-layer matrix-stack interactions is left to future work.
    \item \textbf{Per-benchmark Gram hyperparameters.} As documented in Appendix~\ref{app:impl}, the zero-shot \texttt{Laya-421M} stress-test evaluation uses $\gamma=4.0, \tau_{\text{sim}}=0.8$, though the full milestone-2 sweep across $\tau_{\text{sim}} \in [0, 0.7]$ confirms virtually identical accuracy (55.0--57.0\%) due to Proposition~\ref{prop:margin_bound}'s internal margin inversion.
\end{enumerate}

\section{Related Work}
\label{sec:related}

\subsection{Discrete Choice Theory \& IIA Violations}
\label{sec:related_iia}

Luce's choice axiom \citep{luce1959} states that the ratio of two options' choice probabilities does not depend on which other options are available, and McFadden's conditional logit \citep{mcfadden1974} derives the same property from additive random utility with independent extreme-value errors. A Softmax readout inherits it. Debreu's review of Luce \citep{debreu1960} gave the standard counterexample: adding a near-duplicate of one alternative should not take probability from the others in proportion to the copies, yet the axiom requires that it does. This is the failure we measure as option splitting. Tversky's elimination by aspects \citep{tversky1972} models it directly: choice proceeds by removing options that lack a selected feature, so similar options share features and compete with each other more than with dissimilar ones. These models need the analyst to specify the structure of similarity, through aspects or utility parameters fitted per task. \texttt{GramDiscountedChoiceHead} instead infers redundancy at inference time from the centered similarity of option representations, with no predefined nesting tree, and Theorem~\ref{thm:clone_inv} gives exact invariance when duplicates have identical representations.

\subsection{System-1 Typed Decision Models \& Multiple-Choice Sensitivity}
\label{sec:related_sys1}

Recent single-pass decision models score a fixed set of options with one forward pass and a categorical readout: Narrow Decision Models \citep{ndmblock2026}, Jeva with its Matrix Decision Stack \citep{jeva2026}, the open recipe of \citet{jeff2026} that supplies \texttt{OpenDecisions} and the \texttt{Qwen3-0.6B-RLCD} checkpoint, and the non-autoregressive Laya models \citep{laya2026}. All of them end in an unweighted Softmax over the options. They differ in how options interact before that readout (independent suffixes, joint bidirectional encoding, or a cross-option matrix stack), which is the taxonomy of Section~\ref{sec:arch_families}. We evaluate one model from each family and find that option splitting hurts all three, through different mechanisms (Sections~\ref{sec:results_rq1}--\ref{sec:results_rq2}).

For LLMs answering multiple-choice questions, \citet{pezeshkpour2023} show that reordering the options changes model answers substantially and propose calibration against this order sensitivity. \citet{zheng2024} trace much of the effect to a prior over option-ID tokens and introduce PriDe, which estimates that prior by permuting options on a subset of test samples and removes it at inference. Both address bias over fixed option slots at fixed $K$, using multiple permuted passes or prior calibration. Our setting holds content and order fixed and adds semantically redundant options ($K \to K+m$); we target invariance to that change within a single forward pass.

\subsection{Set-Equivariant Architectures \& Parameter-Efficient Adaptation}
\label{sec:related_sets}

Deep Sets \citep{zaheer2017deepsets} and Set Transformer \citep{lee2019settransformer} make a model invariant or equivariant to the order of set elements, and self-attention \citep{vaswani2017} is permutation-equivariant by construction. Such guarantees concern the permutation $\pi(\mathcal{O})$ at fixed $K$. They do not remove sensitivity to cardinality: a sum-pooled set function counts a duplicated element each time it appears, so a permutation-invariant readout still changes when $m$ copies are added. We target the cross-option attention path directly. Block-diagonal parallel masking removes option-to-option attention while each option keeps access to the question, and we adapt a ModernBERT encoder \citep{modernbert2024} with LoRA \citep{hu2021lora} plus a question-conditioned set adapter (OSA). Together these add 330,881 trainable parameters, 0.08\% of \texttt{Laya-421M}.

\subsection{Confidence Calibration \& Representation Anisotropy}
\label{sec:related_calib}

The Brier score \citep{brier1950} and the 15-bin expected calibration error of \citet{guo2017calibration} are our two calibration metrics. Guo et al.\ also introduce temperature scaling, which divides all logits by one scalar $T$. This preserves the rank order of the logits, so it cannot change which option wins; Table~\ref{tab:stress} shows identical accuracy and decision flip rate for Softmax and temperature scaling at every $m$. A second obstacle is the geometry of contextual embeddings, which occupy a narrow cone \citep{ethayarajh2019anisotropy}. In our data, uncentered distinct options have mean cosine 0.796 (Table~\ref{tab:centering}). Per-question centering removes this offset but also subtracts the shared component of duplicated options, which is why clone similarity falls to 0.193. Rank-preserving temperature scaling and self-subtracting per-query centering both fall short here; we use a fixed global corpus mean estimated once on clean options, combined with logarithmic Gram discounting of the logits.

\section{Conclusion \& Reproducibility Statement}
\label{sec:conclusion}

\subsection{Conclusion}
\label{sec:conclusion_summary}

A single-pass decision interface should give the same answer when an option is duplicated, and standard Softmax readouts do not. On \texttt{Laya-421M}, five paraphrase clones cut accuracy from 88.0\% to 56.0\%. Two interventions address this at different levels. A training-free Gram-discounted head with global corpus mean centering fixes crowding in the readout of independent suffix scorers: on \texttt{Qwen3-0.6B-RLCD} it recovers 24.0 points at $m=3$ and 17.0 points at $m=5$ on \texttt{CLINC150}. Joint bidirectional encoders also need decoupled cross-option attention. Block-diagonal parallel-mask LoRA, a question-conditioned adapter, and pre-attention similarity gating keep 88.0\% clean accuracy, recover 18.0 points at $m=5$, and shrink distractor mass $10.6\times$. The gains are narrower than the headline numbers suggest. The transfer to \texttt{Banking77} is 3.0 points, the 18-point gap has overlapping confidence intervals, and neither intervention helps \texttt{Jeva-4B} or low-accuracy \texttt{OpenDecisions} queries.

\subsection{Reproducibility \& Artifacts Statement}
\label{sec:conclusion_repro}

\vspace{0.3em}
\noindent\fcolorbox{calloutborder}{calloutbg}{%
\parbox{\dimexpr\linewidth-2\fboxsep-2\fboxrule\relax}{%
\small
Code, data, and result logs are publicly archived at \url{https://github.com/shejanahmmed/gram-choice-head}. The repository contains the \texttt{CLINC150} stress test (900 queries), \texttt{Banking77} (200), \texttt{OpenDecisions} (200), the disjoint \texttt{CLINC150} training split (600), \texttt{data/global\_corpus\_mean.pt}, and the JSON logs behind every table in \texttt{runs/}. Trained adapter weights are not included. Adapter training used seeds 42, 123, and 777. Variant B has 330,881 trainable parameters and trained in 274.5 s with 1,896.4 MB peak memory on a single 4 GB GPU (RTX 3050 Laptop). Thirteen automated invariant tests are in \texttt{tests/test\_invariants.py}.
}}
\vspace{0.3em}

\bibliographystyle{plainnat}
\bibliography{references}

\clearpage
\appendix
\renewcommand{\thetable}{A\arabic{table}}
\renewcommand{\theHtable}{A\arabic{table}}
\setcounter{table}{0}

\section{Complete Mathematical Proofs}
\label{app:proofs}

\subsection{Proof of \texorpdfstring{Proposition~\ref{prop:temp}}{Proposition 1} (Temperature Scaling Rank Invariance)}
For any logit vector $z \in \mathbb{R}^K$ and scalar temperature $T > 0$, define $\tilde{z}_i = z_i / T$. Because $x \mapsto x / T$ is strictly increasing on $\mathbb{R}$, $\tilde{z}_i > \tilde{z}_j \iff z_i > z_j$ for all pairs $i, j \in \{1, \dots, K\}$. Because $\exp(\cdot)$ is strictly increasing and the Softmax normalizer $Z(T) = \sum_{k=1}^K \exp(z_k / T) > 0$ is constant across candidates for a given query,
\[
\operatorname{argmax}_{i} P_{\text{soft}}(o_i \mid s, \mathcal{O}; T) = \operatorname{argmax}_{i} \frac{\exp(z_i / T)}{Z(T)} = \operatorname{argmax}_{i} \tilde{z}_i = \operatorname{argmax}_{i} z_i.
\]
Thus scalar temperature scaling leaves top-1 predictions, Top-1 accuracy, and Decision Flip Rate ($\mathbb{I}[\hat{y}_n^{(m)} \ne \hat{y}_n^{(0)}]$) unchanged for every $T > 0$ and every multiplicity $m \ge 0$. \hfill $\square$

\subsection{Proof of \texorpdfstring{Proposition~\ref{prop:dilution}}{Proposition 2} (Softmax Dilution Limit)}
Assume distractor $o_K \ne o^*$ is replicated into $\mathcal{C} = \{o_K, o_{K+1}, \dots, o_{K+m}\}$ ($|\mathcal{C}| = m+1$) with unperturbed logits $z_c = z_K$ for all $c \in \mathcal{C}$ and $z_j$ fixed for $j < K$. Under Eq.~\eqref{eq:softmax}, the Softmax probability of $o^*$ is
\[
P_{\text{soft}}^{(m)}(o^* \mid s, \mathcal{O}^{(m)}) = \frac{\exp(z^*)}{\sum_{j=1}^{K-1} \exp(z_j) + (m+1)\exp(z_K)}.
\]
For $\exp(z_K) > 0$, dividing numerator and denominator by $m+1$ gives $P_{\text{soft}}^{(m)}(o^*) = \mathcal{O}(1/m) \xrightarrow{m \to \infty} 0$, whereas the summed cluster mass satisfies
\[
\sum_{c \in \mathcal{C}} P_{\text{soft}}^{(m)}(c \mid s, \mathcal{O}^{(m)}) = \frac{(m+1)\exp(z_K)}{\sum_{j=1}^{K-1} \exp(z_j) + (m+1)\exp(z_K)} \xrightarrow{m \to \infty} 1.
\]
Comparing total distractor cluster mass against target probability yields
\begin{align*}
\sum_{c \in \mathcal{C}} P_{\text{soft}}^{(m)}(c \mid s, \mathcal{O}^{(m)}) > P_{\text{soft}}^{(m)}(o^* \mid s, \mathcal{O}^{(m)})
&\iff (m+1)\exp(z_K) > \exp(z^*) \\
&\iff m > \exp(z^* - z_K) - 1. \quad \square
\end{align*}

\subsection{Proof of \texorpdfstring{Theorem~\ref{thm:clone_inv}}{Theorem 1} (Exact Clone Invariance \& Mass Conservation)}
Under the hypotheses of Theorem~\ref{thm:clone_inv}, $\hat{\mathbf{e}}_c = \hat{\mathbf{e}}_K$ and $z_c = z_K$ for all $c \in \mathcal{C}$ ($|\mathcal{C}| = m+1$), while every distinct option $i < K$ satisfies $\hat{\mathbf{e}}_i^\top \hat{\mathbf{e}}_j < \tau_{\text{sim}}$ for all $j \ne i$ (including all $c \in \mathcal{C}$, since $\hat{\mathbf{e}}_i^\top \hat{\mathbf{e}}_c = \hat{\mathbf{e}}_i^\top \hat{\mathbf{e}}_K < \tau_{\text{sim}}$). Because $\tau_{\text{sim}} \in (0, 1)$, Definition~\ref{def:gram} gives:
\begin{enumerate}[leftmargin=1.5em]
    \item For any non-cloned option $i < K$, $S_{ii} = 1$ and $S_{ij} = 0$ for all $j \ne i$, so $c_i = \sum_{j=1}^{K+m} S_{ij}^\gamma = 1$.
    \item For any clone $c \in \mathcal{C}$ and $c' \in \mathcal{C}$, $\hat{\mathbf{e}}_c^\top \hat{\mathbf{e}}_{c'} = \|\hat{\mathbf{e}}_K\|_2^2 = 1 \ge \tau_{\text{sim}}$, so $S_{cc'} = 1$ and $S_{cc'}^\gamma = 1$ for any $\gamma \ge 1$, while $S_{ci} = 0$ for all $i < K$. Hence $c_c = \sum_{c' \in \mathcal{C}} 1 = m + 1$.
\end{enumerate}
Substituting $\tilde{z}_i = z_i / T - \log c_i$ into the denominator of Eq.~\eqref{eq:gram_head} for any $T > 0$ gives
\[
Z_{\text{Gram}}^{(m)} = \sum_{i=1}^{K-1} c_i^{-1}\exp(z_i/T) + \sum_{c \in \mathcal{C}} c_c^{-1}\exp(z_c/T) = \sum_{i=1}^{K-1} \exp(z_i/T) + \frac{m+1}{m+1}\exp(z_K/T) = Z_{\text{Gram}}^{(0)}.
\]
Therefore:
\begin{itemize}[leftmargin=1.5em]
    \item \textbf{Axiom 1 (Non-Cloned Invariance):} For every $o_i \notin \mathcal{C}$, $P_{\text{Gram}}^{(m)}(o_i) = \exp(z_i/T) / Z_{\text{Gram}}^{(0)} = P_{\text{Gram}}^{(0)}(o_i)$.
    \item \textbf{Axiom 2 (Cluster Mass Conservation):} $\sum_{c \in \mathcal{C}} P_{\text{Gram}}^{(m)}(c) = (m+1) \cdot \frac{\frac{1}{m+1}\exp(z_K/T)}{Z_{\text{Gram}}^{(0)}} = P_{\text{Gram}}^{(0)}(o_K)$.
    \item \textbf{Axiom 3 (Within-Class Symmetry):} Every $c \in \mathcal{C}$ shares identical discounted logit $\tilde{z}_c = z_K/T - \log(m+1)$, yielding $P_{\text{Gram}}^{(m)}(c) = \frac{1}{m+1} P_{\text{Gram}}^{(0)}(o_K)$. \hfill $\square$
\end{itemize}

\subsection{Derivation of Self-Subtraction Pathology (\texorpdfstring{Eq.~\eqref{eq:sample_mean}}{Eq. (5)})}
Let $\mathbf{a} = \frac{1}{K-1}\sum_{j=1}^{K-1} \mathbf{e}_j$ denote the mean of the $K-1$ non-cloned options and $w_m = \frac{m+1}{K+m}$ the fraction of options in cluster $\mathcal{C}$. Rewriting Eq.~\eqref{eq:sample_mean} gives $\bar{\mathbf{e}}_{\mathcal{O}} = (1 - w_m)\mathbf{a} + w_m \bar{\mathbf{e}}_{\mathcal{C}}$. Decompose each near-clone vector around its cluster centroid as $\mathbf{e}_c = \bar{\mathbf{e}}_{\mathcal{C}} + \boldsymbol{\delta}_c$, where $\sum_{c \in \mathcal{C}} \boldsymbol{\delta}_c = \mathbf{0}$. Subtracting $\bar{\mathbf{e}}_{\mathcal{O}}$ yields
\[
\tilde{\mathbf{e}}_c = \mathbf{e}_c - \bar{\mathbf{e}}_{\mathcal{O}} = (1 - w_m)(\bar{\mathbf{e}}_{\mathcal{C}} - \mathbf{a}) + \boldsymbol{\delta}_c = \frac{K-1}{K+m}(\bar{\mathbf{e}}_{\mathcal{C}} - \mathbf{a}) + \boldsymbol{\delta}_c.
\]
Under exact clones ($\boldsymbol{\delta}_c = \mathbf{0}$), $\|\tilde{\mathbf{e}}_c\|_2 = \frac{K-1}{K+m}\|\bar{\mathbf{e}}_{\mathcal{C}} - \mathbf{a}\|_2 \xrightarrow{m \to \infty} 0$. Under realistic paraphrase variation ($\|\boldsymbol{\delta}_c\|_2 > 0$), as $w_m = \frac{m+1}{K+m} \to 1$, the weight $\frac{K-1}{K+m}$ shrinks the shared cluster displacement $\bar{\mathbf{e}}_{\mathcal{C}} - \mathbf{a}$ below the paraphrase residuals $\boldsymbol{\delta}_c$. After $\ell_2$ normalization, $\hat{\mathbf{e}}_c \approx \boldsymbol{\delta}_c / \|\boldsymbol{\delta}_c\|_2$. Because $\sum_{c \in \mathcal{C}} \boldsymbol{\delta}_c = \mathbf{0}$ implies $\sum_{c \ne c'} \boldsymbol{\delta}_c^\top \boldsymbol{\delta}_{c'} = -\sum_{c \in \mathcal{C}} \|\boldsymbol{\delta}_c\|_2^2 < 0$, the expected pairwise cosine $\hat{\mathbf{e}}_c^\top \hat{\mathbf{e}}_{c'}$ collapses toward $0$ or negative values (0.984 $\to$ 0.193 in Table~\ref{tab:centering}).

\subsection{Proof of \texorpdfstring{Proposition~\ref{prop:margin_bound}}{Proposition 3} (Readout Margin Inversion Bound)}
With $T = 1$, an uncloned target ($c^* = 1 \implies \tilde{z}^* = z^*$), and a cloned distractor with $S_{ij} \in [0, 1]$ restricted to cluster $\mathcal{C}$ ($|\mathcal{C}| = m+1$), effective crowd size satisfies $c_{\text{clone}} \le m+1$. Comparing discounted logits from Eq.~\eqref{eq:gram_head},
\[
\tilde{z}^* > \tilde{z}_{\text{clone}} \iff z^* > z_{\text{clone}} - \log c_{\text{clone}} \iff \Delta z \triangleq z_{\text{clone}} - z^* < \log c_{\text{clone}} \le \log(m+1). \quad \square
\]

\section{Algorithmic Pseudocode \& Training Hyperparameters}
\label{app:impl}

\subsection{PyTorch Reference Pseudocode}
\begin{lstlisting}
# Level-1 Readout: src/gram_choice_head.py (GramDiscountedChoiceHead.forward)
def gram_choice_forward(e, z, valid_mask, global_mean=None, tau_sim=0.5, gamma=2.0, T=1.0):
    z_scaled = z / max(T, 1e-4)
    e_cent = (e - global_mean) * valid_mask.unsqueeze(-1)
    e_hat = F.normalize(e_cent, p=2, dim=-1, eps=1e-8)
    cos = (e_hat @ e_hat.transpose(-1, -2)).clamp(-1.0, 1.0)
    S = torch.where(cos >= tau_sim, cos.clamp_min(0.0), 0.0).pow(gamma)
    eye = torch.eye(z.shape[-1], dtype=torch.bool, device=z.device)
    S = torch.where(eye, 1.0, S) * (valid_mask.unsqueeze(-1) & valid_mask.unsqueeze(-2))
    c = torch.where(valid_mask, S.sum(dim=-1).clamp_min(1.0), 1.0)
    z_tilde = torch.where(valid_mask, z_scaled - torch.log(c), -1e4)
    probs = torch.softmax(z_tilde, dim=-1) * valid_mask
    return z_tilde, probs / probs.sum(dim=-1, keepdim=True).clamp_min(1e-12), c

# Level-2 Structural Decoupling + Gated Routing:
# laya/common.py (parallel_option_masks), src/orthogonal_adapter.py (QuestionConditionedOSA),
# scripts/eval_limitation4_final_sweep.py (DisentangledSimilarityRouter)
def gated_variant_b_forward(opt_ids, attention_mask, e_seq, z_seq, e_par, z_par, h_q, u_tok, u_mean, osa, tau_gate=0.50):
    u_hat = F.normalize(u_tok - u_mean, p=2, dim=-1)
    s_max = (u_hat @ u_hat.t()).masked_select(~torch.eye(u_hat.size(0), dtype=torch.bool, device=u_hat.device)).max()
    if s_max < tau_gate:
        return F.softmax(z_seq, dim=-1), "sequential"
    other_opt = (opt_ids[:, :, None] > 0) & (opt_ids[:, None, :] > 0) & (opt_ids[:, :, None] != opt_ids[:, None, :])
    M_full = attention_mask.bool()[:, None, :] & ~other_opt  # encoder mask with parallel position_ids
    _, probs_par, _, _ = osa(e=e_par, z_base=z_par, h_q=h_q, valid_mask=torch.ones_like(z_par, dtype=torch.bool))
    return probs_par, "variant_b"
\end{lstlisting}

\subsection{Training \& Inference Hyperparameters}

\begin{table}[htbp]
\centering
\caption{Verified hyperparameter configuration across \nolinkurl{src/gram_choice_head.py}, \nolinkurl{src/orthogonal_adapter.py}, \nolinkurl{scripts/train_orthogonal_adapter.py}, \nolinkurl{runs/training_metrics.json}, and \nolinkurl{scripts/eval_limitation4_final_sweep.py}.}
\label{tab:a1}
\footnotesize
\setlength{\tabcolsep}{4pt}
\begin{tabularx}{\textwidth}{L{3.6cm} L{5.1cm} Y}
\toprule
Component / Setting & Primary Value (\texttt{Variant A} / \texttt{Variant B}) & Notes / Source \\
\midrule
Backbone & \texttt{convaiinnovations/laya} (24-layer ModernBERT, $d=1024$) & Frozen in \texttt{Variant A}; upper layers unfrozen via LoRA in \texttt{Variant B} \\
LoRA target layers \& modules & \texttt{layers [20, 21, 22, 23]}, \texttt{attn.Wqkv} & \nolinkurl{scripts/train_orthogonal_adapter.py}; 131,072 params \\
LoRA rank / scaling / dropout & $r=8$, $\alpha_{\text{LoRA}}=16.0$, \texttt{dropout=0.05} & \texttt{LoRALinear} (\nolinkurl{src/orthogonal_adapter.py}) \\
\texttt{QuestionConditionedOSA} & $d=1024$, $r_{\text{OSA}}=64$, $\alpha_{\text{res}}=0.1$ (init), \texttt{dropout=0.05} & Zero-init \texttt{logit\_head}, 199,809 params \\
Total trainable parameters & 199,809 (\texttt{Variant A}) / 330,881 (\texttt{Variant B}) & \nolinkurl{runs/training_metrics.json} ($<0.08\%$ of backbone) \\
Gram readout defaults \& grid & Module default $\tau_{\text{sim}}=0.50, \gamma=2.0$; M2 grid opt $\tau_{\text{sim}}=0.30, \gamma=1.0$ & OSA uses default $(0.50, 2.0)$; script-specific overrides in Section~\ref{sec:eval_protocol} \\
Loss weights & $\lambda_{\text{Brier}}=0.1$, $\lambda_{\text{Ortho}}=0.15$, $\lambda_{\text{Inv}}=0.1$, $\lambda_{\text{Anchor}}=0.5$ & \texttt{Variant A} uses $\lambda_{\text{Anchor}}=0$; \texttt{Variant B} adds clean KL anchor \\
Optimizer \& LR & \texttt{AdamW}, $\text{wd}=10^{-4}$, $\text{lr}_{\text{LoRA}}=3\times 10^{-4}$, $\text{lr}_{\text{OSA}}=10^{-3}$, grad clip 1.0 & Per-example gradient updates (\texttt{train\_items[:200]}/epoch in \texttt{Variant B}) \\
Schedule \& model selection & \texttt{Variant A}: 20 epochs (120 val items); \texttt{Variant B}: 5 epochs (best val loss on \texttt{val\_items[:50]}) & Peak mem: 2501.3 MB (58.7 s, A) / 1896.4 MB (274.5 s, B) \\
3-seed stability protocol & \texttt{seeds = [42, 123, 777]}, 3 epochs, \texttt{weight\_decay=0.01}, final epoch eval & \nolinkurl{scripts/eval_limitation4_final_sweep.py} \\
Disentangled inference gate & Pre-attention \texttt{E\_tok} mean pooling, centered by clean \texttt{m=0} mean, $\tau_{\text{gate}}=0.50$ & \nolinkurl{scripts/eval_limitation4_final_sweep.py} \\
\bottomrule
\end{tabularx}
\end{table}

\section{Benchmark Construction, Disjointness Audits \& Native Prompt Schemas}
\label{app:data}

\subsection{Native Input Prompt Formats}
\begin{itemize}[leftmargin=1.5em]
    \item \textbf{\texttt{Laya-421M}} (\nolinkurl{third_party/laya/laya/common.py}, \nolinkurl{build_sequence} / \nolinkurl{build_head} / \nolinkurl{render_options}):
    Each choice query (\texttt{qtype = 0}) uses \texttt{max\_len=512}, \texttt{head\_max\_len=192}, and per-option token cap \texttt{max\_length=48}:
\begin{lstlisting}[language={}]
[CLS] choice question: <instruction> [SEP] [MASK] <key_0>: <desc_0> [MASK] <key_1>: <desc_1> ... [SEP] <state> [SEP]
\end{lstlisting}
    Sequential inference assigns contiguous \texttt{position\_ids} and full unmasked self-attention; parallel mode (\nolinkurl{parallel_layout} and \nolinkurl{parallel_option_masks}) resets each option span's \texttt{position\_ids} to start at \texttt{markers[0]}, sets \texttt{option\_ids} to $s+1$ for option $s$ (0 for shared context/state), and zeroes cross-option attention entries ($M_{ab} = 0$ when $\text{opt}(a) > 0, \text{opt}(b) > 0, \text{opt}(a) \ne \text{opt}(b)$), preserving \texttt{sliding\_window} offsets via \texttt{position\_ids}.
    \item \textbf{\texttt{Qwen3-0.6B-RLCD}} (\nolinkurl{third_party/eve-rlcd/rlcd/schema.py}, functions \nolinkurl{render_prefix} and \nolinkurl{render_suffix}):\\
    The state prefix is prefilled once into a reusable KV cache (\texttt{SharedPrefixCache}), followed by the choice suffix, with next-token logits read out over \texttt{" A"}, \texttt{" B"}, \dots{} letter token IDs (\nolinkurl{letter_token_ids}):
\begin{lstlisting}[language={}]
# Prefix:
User: Context:
<state>

# Suffix:
Question: <instruction>
Options:
A) <choice_0>
B) <choice_1>
...
Answer with the letter only.
Assistant: The answer is
\end{lstlisting}
    \item \textbf{\texttt{Jeva-4B}} (\nolinkurl{third_party/Jeva/src/jeva/_runtime/collate.py}, \nolinkurl{scripts/eval_limitation2_3.py}):
    State and options are encoded separately via \nolinkurl{state_question_text} and \nolinkurl{option_text} with native key-description rubric pairs \texttt{f"\{k\}: \{crit[k]\}"} before cross-option processing in the 4-layer \texttt{MatrixDecisionStack} ($d=768$):
\begin{lstlisting}[language={}]
# State prompt:
Decision type: choice
Question: <instruction>
State:
<state>
# Option prompt:
Option for choice: <key_i>: <desc_i>
\end{lstlisting}
\end{itemize}

\subsection{Dataset Construction \& Disjointness Verification}
\begingroup\sloppy
\begin{itemize}[leftmargin=1.5em]
    \item \textbf{Controlled Stress Test (\nolinkurl{data/clone_stress_test.jsonl}, $N = 900$):} Built deterministically (\texttt{seed=42}, \nolinkurl{src/clone_benchmark.py}) from 100 base scenarios across 8 support, triage, and security domains, expanded across 9 strictly paired slices (\nolinkurl{control_m0}, \nolinkurl{paraphrase_clone} and \nolinkurl{taxonomic_subsplit} for $m \in \{1,2,3,5\}$). \nolinkurl{scripts/audit_paired_alignment.py} verifies exact 1-to-1 \texttt{base\_id} order equality and 0 symmetric difference across all 9 slices.
    \item \textbf{Disjoint Adapter Training Split (\nolinkurl{data/train_disjoint_600.jsonl}, $N = 600$):} Built in \nolinkurl{scripts/prepare_milestone3_data.py} from \texttt{CLINC150} Travel, Home/Utility, and Kitchen/Dining domains ($K \in [3, 8]$, $m \in \{0,1,2,3\}$; 480 train / 120 val), checked against 18 forbidden test keywords (\texttt{triage}, \nolinkurl{fraud_dispute}, \nolinkurl{failover_replica}, etc.) for 0\% overlap with the stress test. Clean (\texttt{m=0}) options produce \nolinkurl{data/global_corpus_mean.pt} (shape $(1, 1024)$, 5,750 bytes).
    \item \textbf{\texttt{Banking77} (\nolinkurl{data/banking77_eval.jsonl}, $N = 200$):} Built by \nolinkurl{scripts/build_banking77_dataset.py} (\texttt{seed=42}) from \nolinkurl{data/raw_banking77_test.json} (1 query from each of all 77 banking intents + 23 secondary queries = 100 base items paired at \texttt{m=0}, $K=4$ and \texttt{m=5}, $K=9$, with distractors sampled from disjoint intent clusters).
    \item \textbf{\texttt{OpenDecisions} (\nolinkurl{data/opendecisions_eval.jsonl}, $N = 200$):} Built by \nolinkurl{scripts/rebuild_opendecisions_eval.py} across 100 unique questions spanning 11 task families (\texttt{llm\_router}: 20 paired instances; 10 remaining families: 18 paired instances each), paired at \texttt{m=0} ($K \in \{3,4,5,6\}$) and \texttt{m=5} ($K \in \{8,9,10,11\}$).
\end{itemize}
\endgroup

\section{Extended Quantitative Results \& Per-Seed Breakdowns}
\label{app:extended}

Table~\ref{tab:a2} provides full 95\% paired bootstrap confidence intervals across all perturbation levels on the Controlled Stress Test, and Table~\ref{tab:a3} reports individual training runs for \texttt{Variant B} across seeds 42, 123, and 777.

\begin{table}[htbp]
\centering
\caption{Complete 95\% paired bootstrap confidence intervals ($B = 1{,}000$, \texttt{seed = 42}) and calibration metrics on the \texttt{Laya-421M} Controlled Stress Test (100 queries per slice). Sources: \texttt{runs/milestone2\_gram\_head.json} (M2 Best Gram), \texttt{runs/milestone3\_orthogonal\_adapter.json} (Softmax, Temp Scaling, M3 Zero-Shot Gram, Variant A, Variant B), and \texttt{runs/limitation4\_final\_sweep.json} (Gated Variant B). $^\dagger$On \texttt{control\_m0}, \texttt{runs/milestone3\_orthogonal\_adapter.json} logs 0.0000 cluster mass because \texttt{clone\_keys} is empty when $m=0$, whereas \texttt{runs/milestone1\_laya\_baseline.json} logs the uncloned \texttt{target\_distractor} probability (0.0534) and \texttt{runs/limitation4\_final\_sweep.json} (Gated Variant B) aggregates the cached slot entry (0.0187); 100\% of \texttt{control\_m0} queries route to Softmax, so Softmax and Gated Variant B yield identical predictions on all 100 clean queries.}
\label{tab:a2}
\resizebox{\textwidth}{!}{%
\begin{tabular}{llccccc}
\toprule
Method & Slice & Acc (\%) [95\% CI] & DFR (\%) [95\% CI] & ECE & Brier [95\% CI] & Mass \\
\midrule
Softmax & \texttt{control\_m0}$^\dagger$ & 88.0 [81.0, 94.0] & 0.0 [0.0, 0.0] & 0.0899 & 0.1955 [0.1166, 0.2877] & 0.0000 \\
Softmax & \texttt{paraphrase\_m1} & 88.0 [81.0, 94.0] & 4.0 [1.0, 8.0] & 0.1120 & 0.2086 [0.1385, 0.2915] & 0.0756 \\
Softmax & \texttt{paraphrase\_m2} & 87.0 [80.0, 93.0] & 9.0 [4.0, 15.0] & 0.1540 & 0.2677 [0.1898, 0.3600] & 0.1543 \\
Softmax & \texttt{paraphrase\_m3} & 70.0 [60.98, 79.0] & 26.0 [18.0, 35.0] & 0.1347 & 0.3994 [0.3123, 0.5004] & 0.2922 \\
Softmax & \texttt{paraphrase\_m5} & 56.0 [46.0, 66.0] & 40.0 [30.0, 50.0] & 0.1756 & 0.5679 [0.4714, 0.6631] & 0.4852 \\
Softmax & \texttt{taxonomic\_m5} & 77.0 [69.0, 84.0] & 20.0 [13.0, 28.0] & 0.2061 & 0.3707 [0.2992, 0.4491] & 0.3317 \\
Temp Scaling ($T^*=1.11$) & \texttt{paraphrase\_m5} & 56.0 [46.0, 66.0] & 40.0 [30.0, 50.0] & 0.1577 & 0.5700 [0.4807, 0.6599] & 0.4990 \\
M2 Best Gram ($\gamma=1, \tau=0.3$) & \texttt{control\_m0} & 88.0 [81.0, 94.0] & 0.0 [0.0, 0.0] & 0.0828 & 0.1681 [0.1103, 0.2389] & 0.0334 \\
M2 Best Gram ($\gamma=1, \tau=0.3$) & \texttt{paraphrase\_m1} & 88.0 [81.0, 94.0] & 4.0 [1.0, 8.0] & 0.1307 & 0.2222 [0.1636, 0.2914] & 0.1058 \\
M2 Best Gram ($\gamma=1, \tau=0.3$) & \texttt{paraphrase\_m2} & 86.0 [79.0, 92.0] & 10.0 [4.0, 16.02] & 0.1468 & 0.2700 [0.1948, 0.3510] & 0.1269 \\
M2 Best Gram ($\gamma=1, \tau=0.3$) & \texttt{paraphrase\_m3} & 70.0 [61.0, 79.0] & 26.0 [17.98, 35.0] & 0.1250 & 0.3990 [0.3125, 0.4864] & 0.2780 \\
M2 Best Gram ($\gamma=1, \tau=0.3$) & \texttt{paraphrase\_m5} & 57.0 [47.0, 67.0] & 39.0 [29.0, 48.0] & 0.1355 & 0.5609 [0.4649, 0.6552] & 0.4599 \\
M2 Best Gram ($\gamma=1, \tau=0.3$) & \texttt{taxonomic\_m5} & 77.0 [69.0, 85.0] & 20.0 [13.0, 28.0] & 0.1837 & 0.3632 [0.2896, 0.4408] & 0.3122 \\
Zero-Shot Gram ($\gamma=4, \tau=0.8$) & \texttt{paraphrase\_m5} & 56.0 [46.0, 66.0] & 40.0 [30.0, 50.0] & 0.1611 & 0.5725 [0.4638, 0.6822] & 0.4396 \\
Variant A (OSA only) & \texttt{paraphrase\_m5} & 56.0 [46.0, 66.0] & 40.0 [30.0, 50.0] & 0.2540 & 0.6074 [0.4722, 0.7391] & 0.4053 \\
Variant B (Always on) & \texttt{control\_m0} & 78.0 [69.97, 86.0] & 14.0 [8.0, 21.0] & 0.1062 & 0.3177 [0.2182, 0.4328] & 0.0000 \\
Variant B (Always on) & \texttt{paraphrase\_m1} & 81.0 [73.0, 88.0] & 12.0 [6.0, 18.0] & 0.1103 & 0.3083 [0.2117, 0.4235] & 0.0161 \\
Variant B (Always on) & \texttt{paraphrase\_m2} & 80.0 [72.0, 87.02] & 15.0 [9.0, 23.0] & 0.1165 & 0.3182 [0.2177, 0.4391] & 0.0210 \\
Variant B (Always on) & \texttt{paraphrase\_m3} & 80.0 [72.0, 87.02] & 13.0 [7.0, 20.0] & 0.0886 & 0.3252 [0.2225, 0.4437] & 0.0274 \\
Variant B (Always on) & \texttt{paraphrase\_m5} & 74.0 [64.97, 82.0] & 22.0 [15.0, 30.0] & 0.1413 & 0.3730 [0.2641, 0.5023] & 0.0457 \\
Variant B (Always on) & \texttt{taxonomic\_m5} & 74.0 [65.0, 82.0] & 24.0 [16.0, 33.0] & 0.0921 & 0.3602 [0.2659, 0.4650] & 0.1401 \\
Gated Variant B ($\tau_{\text{gate}}=0.50$) & \texttt{control\_m0}$^\dagger$ & 88.0 [81.0, 94.0] & 0.0 [0.0, 0.0] & 0.0899 & 0.1955 [0.1166, 0.2877] & 0.0187 \\
Gated Variant B ($\tau_{\text{gate}}=0.50$) & \texttt{paraphrase\_m5} & 74.0 [64.97, 82.0] & 22.0 [15.0, 30.0] & 0.1413 & 0.3730 [0.2641, 0.5023] & 0.0457 \\
Gated Variant B ($\tau_{\text{gate}}=0.50$) & \texttt{taxonomic\_m5} & 85.0 [77.0, 91.0] & 13.0 [7.0, 20.0] & 0.2068 & 0.3619 [0.2785, 0.4603] & 0.2487 \\
\bottomrule
\end{tabular}%
}
\end{table}

\begin{table}[htbp]
\centering
\caption{Per-seed training stability breakdown for \texttt{Variant B} across seeds 42, 123, and 777 (3 epochs on \texttt{train\_items[:200]}, \texttt{runs/limitation4\_final\_sweep.json}). Mean $\pm$ std omitted where not aggregated in the log.}
\label{tab:a3}
\small
\setlength{\tabcolsep}{5pt}
\begin{tabular}{lcccccc}
\toprule
Seed & \texttt{m0} Acc (\%) & \texttt{m5} Acc (\%) & \texttt{m5} DFR (\%) & \texttt{m5} Mass & \texttt{m5} ECE & \texttt{m5} Brier \\
\midrule
42 & 75.0 & 74.0 & 18.0 & 0.0466 & 0.2026 & 0.4609 \\
123 & 75.0 & 73.0 & 21.0 & 0.0775 & 0.1596 & 0.4027 \\
777 & 73.0 & 69.0 & 23.0 & 0.0565 & 0.1495 & 0.4318 \\
\midrule
\textbf{Mean $\pm$ Std} & \textbf{74.33 $\pm$ 1.15} & \textbf{72.00 $\pm$ 2.65} & \textbf{20.67 $\pm$ 2.52} & \textbf{0.0602 $\pm$ 0.0158} & --- & --- \\
\bottomrule
\end{tabular}
\end{table}

\section{Automated Mathematical Invariant Verification Suite}
\label{app:tests}

All 13 automated unit tests in \nolinkurl{tests/test_invariants.py} verify the formal axioms of the Gram-discounted choice head within $<10^{-6}$ tolerance (13/13 passing via \texttt{python -m pytest tests -q}). Block-diagonal attention mask isolation (\nolinkurl{parallel_option_masks} in \nolinkurl{third_party/laya/laya/common.py}) is verified by mask construction and inspected on 10 query pairs in \nolinkurl{scripts/test_parallel_mask.py} rather than in \nolinkurl{tests/test_invariants.py}.

\begin{table}[htbp]
\centering
\caption{Specification of the 13 \texttt{pytest} unit tests in \nolinkurl{tests/test_invariants.py}.}
\label{tab:a4}
\footnotesize
\begin{tabularx}{\textwidth}{L{0.7cm} L{5.9cm} Y L{1.8cm}}
\toprule
\# & Test Function Name & Verified Mathematical Invariant & Status \\
\midrule
1 & \nolinkurl{test_probability_simplex_and_padding_safety} & Simplex sum $= 1.0$, non-negativity, exact 0.0 on masked slots, valid prob invariance to padding ($<10^{-6}$) & Passed \\
2 & \nolinkurl{test_orthogonal_identity_reduction} & $S = I_K \implies c_i = 1.0$ and exact reduction to standard Softmax ($<10^{-6}$) on QR-orthogonal embeddings & Passed \\
3--11 & \nolinkurl{test_exact_clone_invariance[T-m]} & 9 parametrized cases ($T \in \{0.5, 1.0, 2.0\} \times m_{\text{cluster}} \in \{2, 3, 5\}$): fixture parameter \texttt{m} equals total cluster size $|\mathcal{C}|$ ($c_{\text{clone}} = m_{\text{cluster}}$; Theorem~\ref{thm:clone_inv} uses injected clones $m = |\mathcal{C}| - 1$), Axioms 1--3 ($<10^{-6}$) & Passed (9/9) \\
12 & \nolinkurl{test_option_permutation_invariance} & Candidate permutation equivariance $P(\pi(\mathcal{O})) = \pi(P(\mathcal{O}))$ and identical argmax ($<10^{-6}$) & Passed \\
13 & \nolinkurl{test_global_corpus_mean_centering} & Simplex sum $= 1.0$, non-negativity, and permutation equivariance under fixed \texttt{global\_mean} centering ($<10^{-6}$) & Passed \\
\bottomrule
\end{tabularx}
\end{table}

\end{document}